\documentclass[lettersize,journal]{IEEEtran}
\usepackage{amsmath,amsfonts}
\usepackage{algorithmic}
\usepackage{algorithm}
\usepackage{array}
\usepackage[caption=false,font=normalsize,labelfont=sf,textfont=sf]{subfig}
\usepackage{textcomp}
\usepackage{stfloats}
\usepackage{url}
\usepackage{verbatim}
\usepackage{graphicx}
\usepackage{tabularx}
\usepackage{cite}
\usepackage[hidelinks]{hyperref}
\usepackage{orcidlink} 
\begin{document}

\title{Interruption-Aware Recurrent Reinforcement Learning for UAS Navigation Under Temporary Overrides}

\author{
Christopher~Conrad\,\orcidlink{0000-0003-1077-7086},~\IEEEmembership{Student Member,~IEEE},
Deepak~Panda\,\orcidlink{0009-0003-9925-4448},~\IEEEmembership{Member,~IEEE},
Weisi~Guo\,\orcidlink{0000-0003-3524-3953},~\IEEEmembership{Member,~IEEE},
and Antonios~Tsourdos\,\orcidlink{0000-0002-3966-7633},~\IEEEmembership{Member,~IEEE}%
\thanks{All authors are with the Centre for Assured and Connected Autonomy, Cranfield University,
Bedford, UK. E-mail: \{christopher.conrad, deepak.panda, weisi.guo, antonios.tsourdos\}@cranfield.ac.uk.}%
\thanks{Corresponding author: Christopher Conrad.}%
}

\markboth{PREPRINT - UNDER REVIEW for IEEE Transactions on Intelligent Transportation Systems}%
{Shell \MakeLowercase{\textit{et al.}}: A Sample Article Using IEEEtran.cls for IEEE Journals}

\IEEEpubid{This work has been submitted to the IEEE for possible publication. Copyright may be transferred without notice, after which this version may no longer be accessible.}

\maketitle

\begin{abstract}

Safe autonomous navigation of unmanned aircraft systems (UASs) requires tactical deconfliction to address conflicts that strategic coordination cannot fully prevent. This task is challenging under partial observability, when sensor or traffic data are missing, delayed, or degraded. Geometry-based methods can offer separation assurances but face limitations in dense encounters. Recurrent reinforcement learning (RL) has therefore been proposed to learn navigation policies that integrate observation-action histories to optimise long-term safety-efficiency trade-offs. However, operational safeguards may require a remote pilot, onboard detect-and-avoid (DAA) system, or runtime assurance (RTA) mechanism to temporarily override the learned policy. A research gap remains in understanding how these interruptions affect recurrent policies when the agent regains authority. Benchmark evidence shows that interruptions make task variables less decodable from the recurrent hidden state and increase post-interruption collision risk. The findings suggest that altered recurrent context prolongs post-handback degradation through an action-observation feedback loop, in which changed actions alter the flight path and subsequent recurrent updates. To address this vulnerability, we propose three interruption-aware architectures that condition recurrent processing on interruption context through separate feature streams, dedicated interruption memory, or gated environmental features. Across simulations, all three architectures preserve more decodable task information than a standard recurrent policy. Moreover, the strongest architecture reduces post-interruption collisions by 74\% and interruption-induced mission failures by 69\% relative to the recurrent baseline, without degrading interruption-free performance. These results show that safe interruption does not guarantee safe handback, and support interruption-aware recurrent design for resilient UAS autonomy.
\end{abstract}


\section{Introduction}
\label{section:1_intro}

\IEEEPARstart{U}{nmanned} aircraft system (UAS) traffic management (UTM) frameworks support flight planning, strategic deconfliction and information exchange for UAS navigation \cite{r1,r2}. Nonetheless, the UAS operator or remote pilot remains responsible for flight safety, and strategic coordination cannot eliminate every conflict that develops during a flight. Aircraft may deviate from their submitted intents, surveillance information may be delayed or degraded, and non-cooperative aircraft or other hazards may be absent from networked information. Tactical deconfliction is therefore required to address emerging conflicts and complement strategic UTM services \cite{r3,r13,r28,r37}. Moreover, additional safeguards are required when the tactical navigation algorithm encounters unreliable information or conditions outside its validated operating envelope. An onboard detect-and-avoid (DAA) system, runtime-assurance (RTA) mechanism or remote pilot may temporarily override nominal autonomy to resolve an emerging hazard \cite{r3,r13,r14,r15,r16,r28}. Additionally, interruptions may arise from command-and-control (C2) link disruptions or malicious cyberphysical interference \cite{r50,r51}. The reliability of the nominal algorithm after it regains authority, hereby referred to as handback, is therefore part of the safety problem.

Within this operational framework, tactical deconfliction has traditionally relied on geometry-based methods, including velocity-obstacle and potential-field algorithms \cite{r55,r56}. These approaches often leverage rule-, search- or optimisation-based planning to provide predictable behaviour under explicit assumptions about aircraft dynamics, traffic behaviour and environmental conditions \cite{r4,r5}. Advanced air mobility (AAM), however, will increase the volume, heterogeneity, and autonomy of aircraft in the airspace, creating denser and more uncertain traffic interactions \cite{r1,r27}. These conditions make the behaviour of interacting aircraft harder to model and increase the computational cost of repeatedly solving multi-aircraft deconfliction problems online \cite{r4,r5}.

\IEEEpubidadjcol

Reinforcement learning (RL) offers a complementary approach by learning navigation policies through simulated conflict encounters \cite{r49}. It can optimise safety-efficiency trade-offs while shifting much of the computational burden to offline training \cite{r6,r7,r8,r24}. However, finite sensor coverage, noise, occlusion and degraded traffic information make tactical deconfliction partially observable \cite{r1,r3}. Recurrent RL addresses this limitation by maintaining a hidden state that summarises task-relevant history to inform current decisions \cite{r10,r11,r12}. Notably, Lambrechts et al. \cite{r12} show that recurrent memory encodes information correlated with beliefs over latent task variables, with more informative representations resulting in better task performance. Recurrent memory can therefore be interpreted as a learned, task-dependent information state.

Temporary overrides introduce a distinct challenge for recurrent policies. During an interruption, the policy’s proposed actions are not executed, yet its recurrent state evolves with observations generated by externally selected manoeuvres. This altered context can make task-relevant information less accessible or distort how carried information influences subsequent decisions. Consequently, an interruption that resolves an immediate hazard does not necessarily ensure safe handback.

\subsection{Related Work}

\begin{table*}[!b]
\caption{Summary of existing literature relevant to interrupting recurrent navigation policies.}
\label{table:1_existing_work}
\centering
\footnotesize
\setlength{\tabcolsep}{4pt}
\renewcommand{\arraystretch}{1.15}

\begin{tabularx}{\textwidth}{@{}
    >{\raggedright\arraybackslash}p{0.15\textwidth}
    >{\raggedright\arraybackslash}X
    >{\centering\arraybackslash}p{0.11\textwidth}
    >{\raggedright\arraybackslash}X@{}}
\hline
\textbf{Research area} &
\textbf{Primary focus} &
\textbf{Key references} &
\textbf{Research Gap} \\
\hline

UAS supervision and transport handovers &
Focuses on interruption timing, override conditions, operator readiness, and situational awareness. &
\cite{r3,r13,r14,r15,r16,r17,r18,r26,r28} &
Establishes the operational basis for temporary supervisor overrides, but is primarily concerned with system-level safety and human performance. \\

Recurrent RL for UAS navigation &
Uses recurrent policies for navigation and collision avoidance under partial observability. &
\cite{r11,r36} &
Establishes the relevance of recurrent policies to UAS navigation, but assumes that the nominal policy selects all executed actions. \\

Safe RL and shielded RL &
Incorporates safety constraints into learning and prevents unsafe actions from being executed. &
\cite{r20,r21,r25} &
Learns safe policies and prevents unsafe action execution. Although shielding may replace policy actions, recurrent handback is not considered. \\

Corrigibility and safe interruptibility &
Ensures that agents accept external correction without seeking, resisting, or manipulating interruptions. &
\cite{r19,r31,r32,r33} &
Examines behavioural incentives and the policy learned when interruptions occur during training, but does not consider deployment-time interruptions. \\

Modified-action and human-in-the-loop RL &
Learns when executed actions differ from policy-selected actions, and uses interruptions as corrective feedback. &
\cite{r22,r34} &
Directly represents action substitution, but focuses on its effect on learning rather than recurrent state degradation and post-interruption performance. \\

Recurrent states and modular memory &
Measures information encoded in recurrent states, and structures memory to improve robustness. &
\cite{r12,r23,r29,r35} &
Provides diagnostic and architectural foundations, but does not consider the effect of interruptions. \\

\hline
\end{tabularx}
\end{table*}

Table~\ref{table:1_existing_work} summarises existing research relevant to interrupting recurrent navigation policies. Transportation research examines transitions between automated systems and human decision-makers, focusing on improving override timing and operator readiness \cite{r17,r18,r26}. In UTM and other safety-critical autonomous systems, DAA and RTA research instead addresses better hazard detection, safety monitoring, and recovery behaviour \cite{r3,r13,r14,r15,r16,r28}. Within RL, research primarily addresses the safety of learned behaviour and responsiveness to interruptions. Safe RL incorporates risk measures or safety constraints during training, whereas shielded RL masks or replaces actions that violate safety thresholds \cite{r20,r21,r25}. Corrigibility targets agents that accept or assist external corrections without manipulating the correcting authority \cite{r31,r32,r33}. Specifically, safely interruptible RL addresses how interruptions affect learning, developing algorithms that avoid seeking out or preventing future interruptions \cite{r19}. 

Action-modification research provides the closest formal abstraction of the interruptions considered in this paper. Langlois and Everitt \cite{r34} introduce the modified-action Markov decision process (MAMDP), which distinguishes the action selected by a policy from that ultimately executed following external modifications. Their analysis shows that some agents ignore modifications while others adapt their behaviour to avoid modifications that reduce reward. Human-in-the-loop RL also uses human overrides as demonstrations or feedback to improve policies during training \cite{r22}.

Complementing these studies, recurrent RL has been applied to UAS navigation under partial observability \cite{r11,r36}, while representation analysis examines which latent task variables are decodable from recurrent hidden states \cite{r12,r29}. Influence-aware memory restricts recurrent inputs to variables relevant to unobserved environmental states, reducing interference from irrelevant features \cite{r23}. Recurrent independent mechanisms similarly organise memory into sparsely interacting modules to improve robustness to environmental changes \cite{r35}. These approaches inform the analysis of recurrent representations and the design of structured memory architectures.

\subsection{Research Gap and Contributions}

Existing literature highlights why interruptions are needed, when they should be initiated, and how they affect policy learning. However, a safe interruption does not guarantee safe handback. A research gap remains in understanding how interruptions affect recurrent representations, how they influence post-handback performance, and how they should be accommodated by recurrent policies. We address this gap through the following contributions:

\begin{itemize}
\item Interruption of recurrent UAS navigation is formulated as a handback reliability problem and modelled as a partially observable MAMDP (PO-MAMDP).

\item A paired evaluation framework is developed to characterise the impact of interruptions across different trigger conditions, supervisor behaviours and override durations.

\item Interruptions are shown to reduce task-variable decodability from the recurrent hidden state, degrade agent performance and increase post-handback collision risk. 

\item Three interruption-aware recurrent architectures are proposed that demonstrate increased robustness to interruptions without degrading interruption-free performance.

\end{itemize}

\section{Problem Formulation}
\label{section:2_problem}

\subsection{Operational UTM Context}

\begin{figure}[!b]
\centering
\includegraphics[width=0.49\textwidth]{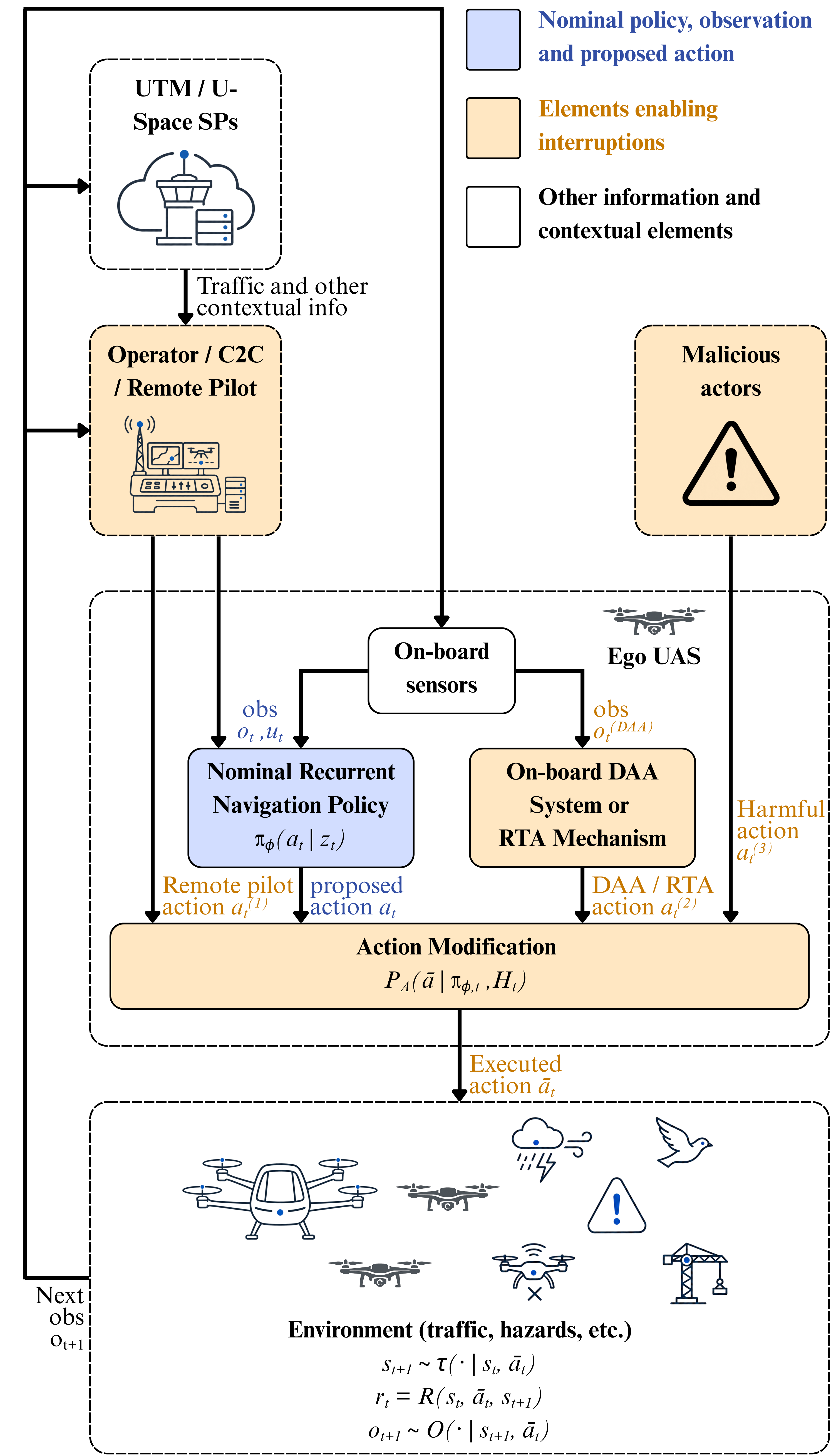}
\caption{Operational context and problem overview for interrupting a recurrent UAS navigation policy.}
\label{fig:1_problem_overview}
\end{figure}

This study focuses on history-dependent tactical navigation under partial observability, as summarised in Table~\ref{tab:1b_operational_scope}. A single operator uses a nominal navigation policy to guide one UAS towards its destination while avoiding surrounding traffic and respecting operational constraints. The policy combines onboard sensor data with traffic and constraint information supplied through UTM services to generate navigation commands. It may run onboard, as illustrated in Fig.~\ref{fig:1_problem_overview}, or in a command-and-control centre (C2C), with commands transmitted over the C2 link. In either configuration, an onboard DAA system, RTA mechanism or remote pilot may temporarily override the policy to resolve emerging threats \cite{r3,r13,r16,r37}. Harmful overrides may also arise from malicious interference \cite{r51}.

\begin{table}[!b]
\caption{Impact of interruptions across UAS navigation functions}
\label{tab:1b_operational_scope}
\centering
\footnotesize
\setlength{\tabcolsep}{4pt}
\renewcommand{\arraystretch}{1.15}

\begin{tabularx}{\columnwidth}{@{}
    >{\raggedright\arraybackslash}p{0.27\columnwidth}
    >{\raggedright\arraybackslash}p{0.28\columnwidth}
    >{\raggedright\arraybackslash}X@{}}
\hline
\textbf{Function and temporal scope} &
\textbf{Representative approaches} &
\textbf{Impact of interruptions and handback} \\
\hline

\textbf{Stabilisation} (milliseconds, low-level control) &
PID, model-based control and physics- informed RL \cite{r52} &
\textbf{Not relevant}. Commands can be recomputed from aircraft state without traffic history.\\

\textbf{Strategic coordination} (pre-flight) &
Scheduling, optimisation and RL/MARL \cite{r53} &
\textbf{Not relevant}. Plans can be revised without interrupting an active navigation policy. \\

\textbf{Short-horizon collision avoidance} (seconds, local manoeuvres) &
Velocity obstacles, potential fields and feedforward RL/MARL \cite{r5} &
\textbf{Not relevant}. A new manoeuvre can be computed from traffic estimates without recurrent context.\\

\textbf{Long-horizon tactical navigation} (minutes-hours, mission execution) &
Recurrent and history-dependent RL \cite{r54} &
\textbf{Primary scope}. Safe handback requires retained traffic and mission context, which is impacted by interruptions.\\

\hline
\end{tabularx}
\end{table}

\subsection{Simulated UAS Deconfliction Task}

In this work, interruption of a nominal recurrent UAS navigation policy is studied using a kinematic simulation designed to isolate the interaction between partial observability, recurrent memory and interruptions. Table~\ref{tab:2_sim_op_mapping} summarises how this environment relates to practical UTM realities.


\begin{table}[!b]
\caption{Operational Interpretation of Simulation Abstractions}
\label{tab:2_sim_op_mapping}
\centering
\footnotesize
\setlength{\tabcolsep}{4pt}
\renewcommand{\arraystretch}{1.15}

\begin{tabularx}{\columnwidth}{@{}
    >{\raggedright\arraybackslash}p{0.27\columnwidth}
    >{\raggedright\arraybackslash}X@{}}
\hline
\textbf{Operational element} &
\textbf{Simulation abstraction} \\
\hline

Tactical UAS navigation &
One ego UAS follows a start-destination mission while avoiding traffic and respecting operational boundaries, representing a vehicle- or operator-side navigation policy. \\

Incomplete traffic information &
Sensor range is limited and a restricted track set is refreshed intermittently, held between updates with an age indicator, and stochastically blanked, representing incomplete or delayed onboard and networked information. \\

Temporary overrides &
An external supervisor replaces the nominal yaw-rate command for a bounded interval, representing a DAA, RTA, remote-pilot or harmful interruption. \\

Handback &
The recurrent policy resumes from the aircraft and memory states reached during an interruption, exposing the post-handback reliability problem. \\

Operational constraints and objectives &
Boundary clamping approximates an operational-volume or geozone limit, and the reward defines safety and efficiency objectives for a finite learning task. \\

\hline
\end{tabularx}
\end{table}

\paragraph{World and ego dynamics}

A single UAS must navigate to its destination within a bounded airspace volume, while avoiding multiple moving intruders. The ego aircraft travels at constant ground speed in a horizontal plane. At each timestep, the navigation policy proposes a normalised yaw-rate command $a_t\in[-1,1]$, but during an interruption the action proposed by the supervisor is executed instead. The ego dynamics are hence given by \eqref{eq:1_ego_dynamics}.
\begin{align}
\psi_{t+1}
&=\psi_t+\dot{\psi}_{\max}\bar a_t\Delta t,\nonumber\\
\mathbf{p}_{t+1}
&=\mathbf{p}_t+v_a
\begin{bmatrix}
\cos\psi_{t+1} & \sin\psi_{t+1}
\end{bmatrix}^{\!\top}\Delta t,
\label{eq:1_ego_dynamics}
\end{align}

\noindent where $\bar a_t$ is the executed action; $\dot{\psi}_{\max}$ is the maximum yaw rate; and $\mathbf{p}_t=[x_t,y_t]^\top$, $\psi_t$ and $v_a$ are the ego position, heading and ground speed, respectively.

Start and destination positions are randomised in each episode. Moreover, surrounding traffic includes both intruders whose initial trajectories intersect the nominal start-destination path and intruders sampled elsewhere in the operating region, producing a range of direct and indirect encounters. Intruder headings are perturbed stochastically, preventing future encounter geometry from being inferred from a fixed trajectory model. An episode succeeds when the ego UAS reaches the destination and terminates unsuccessfully upon collision or when the mission horizon elapses. The ego position is also clamped at the operating-region boundary, analogous to an operational-volume or geozone limit \cite{r37}. Exact supervisor behaviours, trigger conditions, override durations, intruder speeds, traffic densities, and sampling distributions are reported in Section~\ref{section:4_evaluation}.

\paragraph{Observations and partial observability}

The environmental observation contains the information listed in \eqref{eq:2_env_obs}.
\begin{equation}
o_t=\big[
\underbrace{\rho_{1:K,t},\beta_{1:K,t},\tau_{1:K,t}}
_{\text{traffic tracks}};
\underbrace{\mathbf{d}^{\mathrm{bnd}}_t}_{\text{boundaries}};\underbrace{d^g_t,\beta^g_t}_{\text{destination}};
\underbrace{\dot\psi_t,\bar a_{t-2}}_{\text{ego motion}}
\big],
\label{eq:2_env_obs}
\end{equation}

\noindent where $K$ is the maximum number of intruders the ego can observe simultaneously within a maximum sensing range, each reporting range $\rho_{i,t}$, relative bearing $\beta_{i,t}$, and measurement age $\tau_{i,t}$. Additionally, $\mathbf{d}^{\mathrm{bnd}}_t$ contains distances to the operating-region boundaries; $d^g_t$ and $\beta^g_t$ are the observed destination range and bearing; and $\dot\psi_t=\dot\psi_{\max}\bar a_{t-1}$ and $\bar a_{t-2}$ reflect past actions of the UAS.

Partial observability is further imposed through temporally degraded information. At the beginning of each episode, a common traffic-track refresh period is sampled. All $K$ slots are refreshed together at this period and held between refreshes with an increasing age indicator. A separate dropout process may stochastically blank the track information, and the destination channel is also periodically unavailable. Section~\ref{section:4_evaluation} reports the exact degradation schedules used.

\paragraph{Reward}

The agent is rewarded for reaching its destination, separating from other traffic, and avoiding unnecessarily abrupt or prolonged manoeuvres, according to \eqref{eq:3_reward}.
\begin{align}
r_t={}&k_p(D^g_{t-1}-D^g_t)
-k_c[d_{\mathrm{safe}}-d_t^{\min}]_+
\notag\\
&-k_{\Delta a}|\bar a_t-\bar a_{t-1}|-c_{\mathrm{time}}
\notag\\
&+B_g\mathbb{I}\{\mathrm{goal}_t\}
-P_c\mathbb{I}\{\mathrm{collision}_t\}
-P_b\mathbb{I}\{\mathrm{bnd}_t\},
\label{eq:3_reward}
\end{align}

\noindent where $D_t^g$ is the distance to the goal; $d_t^{\min}$ is the distance to the nearest intruder; $d_{\mathrm{safe}}$ is a prescribed separation distance; and $c_{\mathrm{time}}$ is a per-step time penalty. The operator $[x]_+=\max(0,x)$ activates the separation penalty only when $d_t^{\min}<d_{\mathrm{safe}}$, and $\mathbb{I}\{\cdot\}$ is an indicator function, equal to one when the specified event occurs and zero otherwise. Additionally, $k_p$, $k_c$, $k_{\Delta a}$, $B_g$, $P_c$, and $P_b$ are configurable weights defined in Section~\ref{section:4_evaluation} to reflect different reward trade-offs. All terms are evaluated from the simulator state and executed actions, even when the corresponding information is degraded or unavailable in the policy observation. This ensures that observation degradation does not alter the task objective.

\subsection{UAS Navigation as a POMDP}

When the state can be fully observed, UAS navigation is modelled as an MDP $
\mathcal{M}^{\mathrm{FO}}=
(\mathcal{S},\mathcal{A},\mathcal{T},\mathcal{R},\rho_0,\gamma),$
where $\mathcal{S}$ is the state space, $\mathcal{A}$ is the action space, $\mathcal{T}(s' \mid s,a)$ is the state-transition kernel, $\mathcal{R}(s,a,s')$ is the reward function, $\rho_0$ is the initial-state distribution, and $\gamma \in [0,1)$ is the discount factor. In practice, however, some information is unavailable, delayed or degraded \cite{r1,r3}. Consequently, UAS navigation is modelled as a partially observable MDP (POMDP) $\mathcal{M} =(\mathcal{S},\mathcal{A},\mathcal{T},\mathcal{R},\Omega,\mathcal{O},\rho_0,\gamma),$ which extends the MDP with observation space $\Omega$ and an observation kernel $\mathcal{O}(o \mid s', a)$ that gives the probability of observing $o$ when action $a$ leads to state $s'$ \cite{r9}. Since the policy does not observe $s_t$ directly, it must leverage the observation-action history, which under nominal execution is given by \eqref{eq:4_pomdp_history}.
\begin{equation}
\label{eq:4_pomdp_history}
H_t=(o_0,a_0,o_1,a_1,\ldots,a_{t-1},o_t).
\end{equation}

An exact POMDP solution may condition on the belief state $b_t(s)=\Pr(s_t=s\mid H_t)$. However, analytically computing this belief is generally intractable in high-dimensional, continuous or unknown environments \cite{r9,r47}. A recurrent policy therefore compresses its available observation history into a learned hidden state according to \eqref{eq:5_recurrent_policy_overview}.
\begin{equation}
\pi_{\phi,t}(\cdot)=\pi_\phi(\cdot\mid z_t),
\qquad
z_t=F_\phi(z_{t-1},o_t),
\label{eq:5_recurrent_policy_overview}
\end{equation}

\noindent where $z_t$ denotes the recurrent hidden state and $\phi$ denotes the policy parameters. The nominal policy action is sampled as $a_t\sim\pi_{\phi,t}$. Consequently, for an episode of horizon $N$, the policy objective is given by \eqref{eq:6_pomdp_obj}.
\begin{align}
J_{\mathcal{M}}(\phi)&=
\mathbb{E}_{\tau_{nom}\sim p_{\phi,\mathcal{M}}}
\left[
\sum_{t=0}^{N-1}\gamma^t r_t
\right],\nonumber\\
r_t&=\mathcal{R}(s_t,a_t,s_{t+1}),
\label{eq:6_pomdp_obj}
\end{align}

\noindent where $\tau_{nom}$ is the resulting nominal trajectory and $p_{\phi,\mathcal{M}}$ is the trajectory distribution induced jointly by the policy and POMDP dynamics.

\subsection{Interrupted UAS Navigation as a PO-MAMDP}

A standard MDP assumes that the actions proposed by the nominal policy are always executed. Langlois and Everitt \cite{r34} relax this assumption through the modified-action MDP (MAMDP), which augments an MDP with an action-selection function $P_A\left(a\mid\pi,s\right)$, that maps a policy and state to a distribution over the actions actually executed. They term the unmodified policy the \emph{virtual policy} and the resulting distribution over executed actions the \emph{empirical policy}.

We extend this notion by defining a PO-MAMDP in \eqref{eq:7_pomamdp_definition},
\begin{equation}
\widetilde{\mathcal{M}}=(\mathcal{M},P_A),
\label{eq:7_pomamdp_definition}
\end{equation}

\noindent with the resulting empirical policy given by \eqref{eq:8_mamdp_empirical_policy}.
\begin{equation}
\dot{\pi}_{\phi,t}(\bar a\mid H_t)
:=
P_A(\bar a\mid\pi_{\phi,t},H_t),
\label{eq:8_mamdp_empirical_policy}
\end{equation}

\noindent where $\bar a$ denotes an executed action and $\dot{\pi}_{\phi,t}$ is the empirical policy induced by action modification. The ordinary POMDP can be recovered when the virtual policy is applied without modification, according to \eqref{eq:9_unmodified_action_selection}.
\begin{equation}
P_A^{\mathrm{POMDP}}(\bar a\mid\pi_{\phi,t},H_t)
=\pi_{\phi,t}(\bar a).
\label{eq:9_unmodified_action_selection}
\end{equation}

For simplicity, we consider $H_t\in\mathcal{H}$ to denote the complete joint interaction history available at $t$ for both the nominal policy and supervisors. Nonetheless, a policy or function conditioned on $H_t$ is understood to only use the information available to it. For example, an onboard DAA function may receive local surveillance measurements of hazards that are absent from the networked information used by a ground-based navigation policy. Conversely, a remote pilot may have access to operational intents, airspace constraints, or supplementary traffic information that is unavailable onboard \cite{r1,r3}.

\subsection{Interruptions in the UAS Navigation PO-MAMDP}

Orseau and Armstrong \cite{r19} formally define an interruption scheme by the triplet $\langle I,\theta,\pi^{\mathrm{INT}}\rangle$. The initiation function $I(H_t)\in[0,1]$ assesses whether an interruption should begin after the current history, $\theta_t\in[0,1]$ scales the probability that the interruption policy is applied when initiated, and $\pi^{\mathrm{INT}}$ specifies the actions selected during the interruption. The authors further introduce an interruption operator $INT^\theta:\Pi \rightarrow \Pi$ that turns a policy $\pi$ into an interruptible policy $INT^\theta\left(\pi\right)$ given by \eqref{eq:10_int_operator}, where the agent follows $\pi^{INT}$ when it is interrupted and $\pi$ otherwise.
\begin{align}
\mathrm{INT}^{\theta}(\pi)(a\mid H_t)
={}&\theta_t I(H_t)\pi^{\mathrm{INT}}(a\mid H_t)
\notag\\
&+\bigl[1-\theta_t I(H_t)\bigr]\pi(a\mid H_t).
\label{eq:10_int_operator}
\end{align}

We extend the definition of an interruption to include a termination function $\zeta_k(H_t,\ell)\in[0,1]$, giving the probability that supervisor $k$ hands decision authority back to the nominal policy after $\ell$ active steps. Moreover, we assume that an applied interruption will always trigger successfully in deployment, corresponding to $\theta_t=1$. $\theta_t$ is hence omitted from subsequent equations for simplicity. With this definition, overrides can be defined as interruptions that modify nominal policy actions within the PO-MAMDP. Let $m_t\in\{0,1,\ldots,K_{\mathrm{sup}}\}$ identify the source of the executed action of the empirical policy, where $m_t=0$ denotes nominal execution and $m_t=k>0$ denotes supervisor $k$. When no interruption is active, the probability that supervisor $k$ initiates an override is given by \eqref{eq:11_interruption_initiation}.
\begin{equation}
\begin{aligned}
&\Pr(m_t=k\mid m_{t-1}=0,H_t)\\
&\quad=I(H_t)q_t(k\mid H_t),
\qquad k=1,\ldots,K_{\mathrm{sup}},
\end{aligned}
\label{eq:11_interruption_initiation}
\end{equation}

\noindent where $q_t(k\mid H_t)$ is the conditional distribution used to select the supervisor and $K_{sup}$ is the number of distinct supervisors considered. Similarly, the probability of a supervisor handing back control to the nominal policy is given by \eqref{eq:12_handback_probability}.
\begin{equation}
\Pr(m_{t+1}=0\mid m_t=k,H_t,\ell)
=\zeta_k(H_t,\ell).
\label{eq:12_handback_probability}
\end{equation}

A fixed-duration interruption is obtained by making handback deterministic after a sampled duration $d_j$, and a scheduled interruption is obtained by concentrating $I$ at an onset $t_j^{\mathrm{on}}$. In general, an interruption $j$ is therefore the interval \eqref{eq:13_interruption_interval},
\begin{equation}
\mathcal{I}_j=
\{t_j^{\mathrm{on}},\ldots,t_j^{\mathrm{HB}}-1\},
\qquad
t_j^{\mathrm{HB}}=t_j^{\mathrm{on}}+d_j,
\label{eq:13_interruption_interval}
\end{equation}
during which $m_t=k_j>0$. Handback occurs at $t_j^{\mathrm{HB}}$, when the nominal policy again determines the executed action.

The source $m_t$ forms part of the joint history $H_t$, and defining $P_{A,t}(\bar a)\equiv P_A(\bar a\mid\pi_{\phi,t},H_t)$ gives \eqref{eq:14_mamdp_action_selection}.
\begin{equation}
P_{A,t}(\bar a)=
\begin{cases}
\pi_{\phi,t}(\bar a), & m_t=0,\\
\pi_k^{\mathrm{INT}}(\bar a\mid H_t), & m_t=k>0.
\end{cases}
\label{eq:14_mamdp_action_selection}
\end{equation}

\noindent Equivalently, if $a_t\sim\pi_{\phi,t}$ is the nominal proposal and supervisor $k$ supplies an explicit command $a_t^{(k)}$, the executed action is given by \eqref{eq:15_action_substitution}.
\begin{equation}
\bar a_t=
\begin{cases}
a_t, & m_t=0,\\
a_t^{(k)}, & m_t=k>0.
\end{cases}
\label{eq:15_action_substitution}
\end{equation}




\section{Interruption-Aware Recurrent RL}
\label{section:3_proposal}

\subsection{LSTM-based PPO as a Recurrent Baseline}

Several RL algorithms have been applied to UAS navigation under partial observability \cite{r11,r36}. Among policy-gradient methods, proximal policy optimisation (PPO) is particularly suitable because it enables several minibatch updates from each collected rollout and discourages excessively large policy changes through a clipped surrogate objective \cite{r38}. Unlike value-based methods such as deep Q-learning, PPO can represent continuous actions without discretising the action space \cite{r44}. It also directly represents a stochastic policy, unlike deep deterministic policy gradient (DDPG) methods, and avoids the complex constrained optimisation required by trust-region policy optimisation (TRPO) \cite{r45,r46}. A recurrent PPO agent therefore serves as the baseline policy for this work, to optimise a stochastic distribution over the continuous yaw-rate commands of the ego UAS described in Section~\ref{section:2_problem}.

For each rollout, let $\phi_{\mathrm{old}}$ denote the actor parameters used to collect the data. PPO measures the change in probability assigned to a sampled action using the likelihood ratio in \eqref{eq:18_ppo_ratio}.
\begin{equation}
\begin{aligned}
\eta_t(\phi)
&=
\frac{\pi_\phi(a_t\mid z_t^\phi)}
{\pi_{\phi_{\mathrm{old}}}(a_t\mid z_t^{\mathrm{old}})},\\
\bar{\eta}_t(\phi)
&=
\operatorname{clip}\!\left(
\eta_t(\phi),1-\epsilon,1+\epsilon
\right),
\end{aligned}
\label{eq:18_ppo_ratio}
\end{equation}
where $\eta_t$ is the likelihood ratio, $\bar{\eta}_t$ is its clipped value, $\epsilon$ is the clipping threshold, and $z_t^\phi$ and $z_t^{\mathrm{old}}$ are the recurrent states produced by the updated and rollout policies, respectively. Given an advantage estimate $\widehat A_t$, commonly calculated using generalised advantage estimation (GAE) \cite{r39}, the actor maximises \eqref{eq:19_ppo_clipped_objective}.
\begin{equation}
L_{\pi}^{\mathrm{clip}}(\phi)=
\mathbb{E}_t\!\left[
\min\!\left(
\eta_t(\phi)\widehat A_t,
\bar{\eta}_t(\phi)\widehat A_t
\right)
\right].
\label{eq:19_ppo_clipped_objective}
\end{equation}
In the actor-critic formulation, the actor $\pi_\phi$ defines the action distribution, while the critic $V_\psi$ estimates the expected discounted return from the current state. The resulting advantage estimate indicates whether an action performed better or worse than the critic predicted and therefore guides the actor update. The complete PPO loss additionally contains a value-prediction loss for the critic and an entropy term that discourages premature policy convergence \cite{r38}.

Recurrent policies address partial observability by compressing the observation-action history into a learned information state that can encode task-relevant belief information \cite{r12,r23}. In this work, long short-term memory (LSTM) networks are used because their gated cell structure supports information retention over longer temporal dependencies than an ungated recurrent neural network \cite{r40}. Although gated recurrent units (GRUs) offer a simpler gated alternative, LSTMs maintain a distinct cell state that provides an explicit pathway for longer-term information retention \cite{r41}. This separation is particularly useful for analysing how task-relevant memory changes during an interruption.

For recurrent input $x_t$, an LSTM updates its cell state $c_t$ and hidden state $z_t$ according to \eqref{eq:20_lstm_update}.
\begin{equation}
\begin{aligned}
g_t^{\mathrm f}
 &=\sigma(W_{\mathrm f}x_t+U_{\mathrm f}z_{t-1}+b_{\mathrm f}),\\
g_t^{\mathrm i}
 &=\sigma(W_{\mathrm i}x_t+U_{\mathrm i}z_{t-1}+b_{\mathrm i}),\\
g_t^{\mathrm o}
 &=\sigma(W_{\mathrm o}x_t+U_{\mathrm o}z_{t-1}+b_{\mathrm o}),\\
\widetilde c_t
 &=\tanh(W_{\mathrm c}x_t+U_{\mathrm c}z_{t-1}+b_{\mathrm c}),\\
c_t
 &=g_t^{\mathrm f}\odot c_{t-1}
   +g_t^{\mathrm i}\odot\widetilde c_t,\\
z_t
 &=g_t^{\mathrm o}\odot\tanh(c_t),
\end{aligned}
\label{eq:20_lstm_update}
\end{equation}
where $\sigma$ denotes the logistic sigmoid; $\odot$ denotes element-wise multiplication; $g_t^{\mathrm f}$, $g_t^{\mathrm i}$, and $g_t^{\mathrm o}$ denote the forget-, input-, and output-gate values, respectively; and $W$, $U$, and $b$ denote learned weights and biases.

The recurrent baseline is therefore an LSTM-based PPO agent with separate actor and critic recurrent states. Nonetheless, the deployment-time failure mechanism should apply more generally to interrupted history-dependent policies. Similarly, the proposed interruption-aware architectures can be adapted to other recurrent networks, although the masked actor objective introduced later in this section is formulated specifically for PPO.


\subsection{Interruption-Aware Observations}

The environmental observation in \eqref{eq:2_env_obs} contains past executed actions but does not identify whether those actions were selected by the nominal policy. We therefore augment this observation by the input $u_t$ in \eqref{eq:21_intervention_observation}.
\begin{equation}
u_t=
\bigl[a_{t-1},\lambda_t\bigr],
\qquad
\lambda_t=\mathbb{I}\{m_{t-1}>0\},
\label{eq:21_intervention_observation}
\end{equation}

\noindent where $a_{t-1}$ is the action proposed by the nominal policy and $\mathbb{I}(\cdot)$ is the indicator function. $\lambda_t$ is a binary flag that indicates whether the preceding action was replaced by a supervisor. This reflects the operational reality that navigation policy may not know which supervisor intervened or why, but the vehicle system can indicate that its nominal proposal was replaced by an externally selected action. All proposed architectures receive both $o_t$ and $u_t$, but differ in how these input streams are encoded, combined, and incorporated into recurrent memory.

\subsection{Interruption-Aware Recurrent Architectures}

An interruption-aware policy receives two qualitatively different input streams. The environmental observation $o_t$ describes the aircraft, surrounding traffic, destination, and environment. Although its measurements may be intermittent or degraded, the underlying physical state generally evolves continuously over time. Conversely, the interruption input $u_t$ changes abruptly at an interruption trigger and upon handback. The recurrent agent must therefore construct a task-relevant belief for the evolving physical environment while also representing sparse changes in decision authority.

\begin{figure*}[!b]
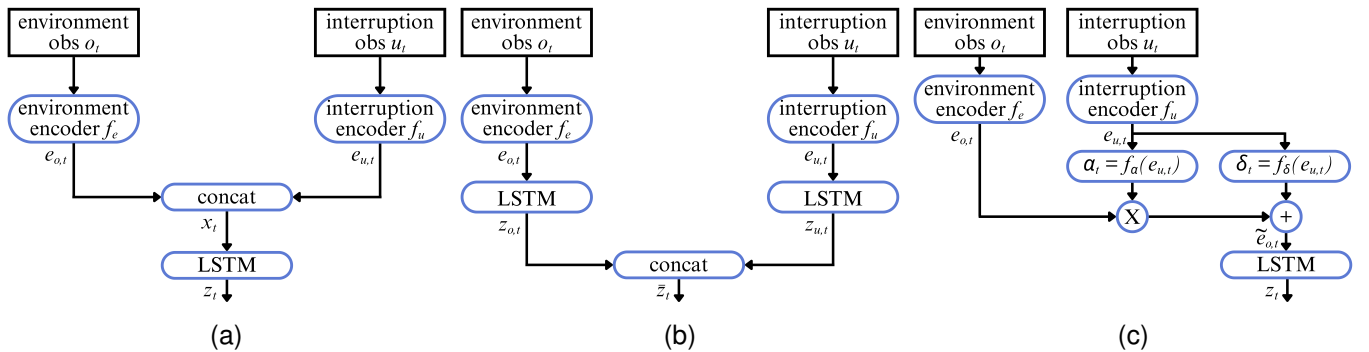

\centering
\subfloat[]{%
\includegraphics[height=1.6in]{figures/fig4a_df.png}%
\label{fig:4a_df}}
\hfil
\subfloat[]{%
\includegraphics[height=1.6in]{figures/fig4b_dm.png}%
\label{fig:4b_dm}}
\hfil
\subfloat[]{%
\includegraphics[height=1.6in]{figures/fig4c_g.png}%
\label{fig:4c_g}}
\caption{Overview of proposed interruption-aware recurrent architectures: (a) Dual-feature architecture. (b) Dual-memory architecture. (c) Gated architecture.}
\label{fig:4_interruption_aware_architectures}
\end{figure*}

Three interruption-aware recurrent architectures are proposed to explicitly encode this distinction, as summarised in Fig.~\ref{fig:4_interruption_aware_architectures}. All three process the environmental and interruption inputs through separate learned encoders according to \eqref{eq:22_dual_encoders}.
\begin{equation}
e_{o,t}=f_e(o_t)\in\mathbb{R}^{d_o},
\qquad
e_{u,t}=f_u(u_t)\in\mathbb{R}^{d_u},
\label{eq:22_dual_encoders}
\end{equation}

\noindent where $f_e$ and $f_u$ are the environmental and interruption branches of a feature extractor shared by the actor and critic. The resulting features $e_{o,t}$ and $e_{u,t}$ provide separate representations of the physical environment and interruption context. The architectures differ in where and how these representations are combined within the actor. To keep value estimation consistent, the critic always retains a single LSTM.

Algorithm~\ref{alg:1_architecture_forward} summarises one actor-critic forward pass for each architecture. Each recurrent state is written as $s_t=(c_t,z_t)$, where $z_t$ is its hidden-state component. The variable $y_t$ denotes the representation supplied to the policy head, while $x_t^V$ denotes the input supplied to the critic. The critic hidden state $z_t^V$ is processed by the post-recurrent layer $g_V$ and the scalar value head $V_\psi$. Its output $\widehat v_t$, estimates the expected discounted return from the current interaction history under the learned policy and interruption process.

\subsubsection{Dual-feature architecture}

The dual-feature architecture provides the least restrictive separation between environmental and interruption information. The two inputs are encoded independently, concatenated, and supplied to a single actor LSTM. Let
$s_t^{\mathrm{DF}}=(c_t^{\mathrm{DF}},z_t^{\mathrm{DF}})$
denote its cell and hidden states. The actor is then defined by \eqref{eq:23_dual_feature_policy}.
\begin{align}
x_t^{\mathrm{DF}}
&=[e_{o,t};e_{u,t}],
\notag\\
s_t^{\mathrm{DF}}
&=\operatorname{LSTM}_{\mathrm{DF}}
  (x_t^{\mathrm{DF}},s_{t-1}^{\mathrm{DF}}),
\notag\\
\pi_{\phi,t}^{\mathrm{DF}}(\cdot)
&=\pi_{\phi}^{\mathrm{DF}}
  \bigl(\cdot\mid g_{\mathrm{DF}}(z_t^{\mathrm{DF}})\bigr),
\label{eq:23_dual_feature_policy}
\end{align}

\noindent where, $x_t^{\mathrm{DF}}$ is the concatenated feature vector, $g_{\mathrm{DF}}$ is the post-LSTM feed-forward layer, and $\pi_{\phi,t}^{\mathrm{DF}}$ is the resulting action distribution. Separate feature extraction makes interruption status explicit without prescribing how it must be stored. Consequently, both feature streams update the same recurrent hidden state, allowing environmental and interruption information to interact, and potentially interfere, within the shared memory. The critic receives the same feature vector but processes it using its own LSTM and value head.

\subsubsection{Dual-memory architecture}

The dual-memory architecture imposes a stronger separation by assigning the environmental and interruption features to independent actor LSTMs. Let
$s_{o,t}^{\mathrm{DM}}=(c_{o,t}^{\mathrm{DM}},z_{o,t}^{\mathrm{DM}})$
and
$s_{u,t}^{\mathrm{DM}}=(c_{u,t}^{\mathrm{DM}},z_{u,t}^{\mathrm{DM}})$
denote their respective recurrent states. The actor is defined by \eqref{eq:24_dual_memory_policy}.
\begin{align}
s_{o,t}^{\mathrm{DM}}
&=\operatorname{LSTM}_{o}
  (e_{o,t},s_{o,t-1}^{\mathrm{DM}}),
\notag\\
s_{u,t}^{\mathrm{DM}}
&=\operatorname{LSTM}_{u}
  (e_{u,t},s_{u,t-1}^{\mathrm{DM}}),
\notag\\
y_t^{\mathrm{DM}}
&=g_{\mathrm{DM}}
  ([z_{o,t}^{\mathrm{DM}};z_{u,t}^{\mathrm{DM}}]),
\notag\\
\pi_{\phi,t}^{\mathrm{DM}}(\cdot)
&=\pi_{\phi}^{\mathrm{DM}}
  (\cdot\mid y_t^{\mathrm{DM}}),
\label{eq:24_dual_memory_policy}
\end{align}

\noindent where, the environmental LSTM maintains a history-dependent representation of the physical encounter, while the interruption LSTM tracks the nominal proposal and interruption context. Their hidden states are combined only after recurrence by the learned fusion layer $g_{\mathrm{DM}}$, producing the policy feature $y_t^{\mathrm{DM}}$. This separation prevents the interruption features from directly updating the environmental recurrent state, although $o_t$ may still contain observations generated by supervisor-selected manoeuvres. The critic receives the concatenated encoder output $[e_{o,t};e_{u,t}]$ but processes it using its own single LSTM and value head. 

\subsubsection{Gated architecture}

The gated architecture combines the two input streams through feature-wise linear modulation (FiLM) \cite{r43}. Rather than supplying the interruption features directly to the recurrent layer, FiLM uses them to scale and shift each environmental feature according to \eqref{eq:25_film_modulation}.
\begin{align}
\alpha_t
&=f_{\alpha}(e_{u,t}),
\notag\\
\delta_t
&=f_{\delta}(e_{u,t}),
\notag\\
\widetilde e_{o,t}
&=\alpha_t\odot e_{o,t}+\delta_t,
\label{eq:25_film_modulation}
\end{align}

\noindent where $f_{\alpha}$ and $f_{\delta}$ are learned mappings from the interruption features to feature-wise scale and shift vectors, respectively; $\odot$ denotes element-wise multiplication; and $\widetilde e_{o,t}$ is the modulated environmental representation. Let
$s_t^{\mathrm{G}}=(c_t^{\mathrm{G}},z_t^{\mathrm{G}})$
denote the state of the actor LSTM. The actor is then defined by \eqref{eq:26_feature_gated_policy}.
\begin{align}
s_t^{\mathrm{G}}
&=\operatorname{LSTM}_{\mathrm{G}}
  (\widetilde e_{o,t},s_{t-1}^{\mathrm{G}}),
\notag\\
\pi_{\phi,t}^{\mathrm{G}}(\cdot)
&=\pi_{\phi}^{\mathrm{G}}
  \bigl(\cdot\mid g_{\mathrm{G}}(z_t^{\mathrm{G}})\bigr).
\label{eq:26_feature_gated_policy}
\end{align}

\noindent This architecture allows the interruption context to determine which environmental features are emphasised, attenuated, or shifted before they update recurrent memory. It therefore provides stronger interaction between the two input streams than dual memory, while avoiding their direct concatenation within a shared recurrent update. The critic receives the same modulated environmental representation $\widetilde e_{o,t}$ but processes it using its own LSTM and value head. 

\begin{algorithm}[!b]
\caption{Forward pass of interruption-aware architectures}
\label{alg:1_architecture_forward}
\footnotesize
\begin{algorithmic}[1]
\REQUIRE $o_t$, $a_{t-1}$, $\lambda_t$, previous recurrent states and architecture $A$

\STATE Form interruption input:
       $u_t\gets[a_{t-1},\lambda_t]$
\STATE Encode context:
       $e_{o,t}\gets f_e(o_t), e_{u,t}\gets f_u(u_t)$

\IF{$A=\textsc{Dual Feature}$}
    \STATE Concatenate features:
           $x_t^{\mathrm{DF}}\gets[e_{o,t};e_{u,t}]$
    \STATE Update shared memory $s_t^{\mathrm{DF}}$
           using $x_t^{\mathrm{DF}}$
    \STATE Extract policy features:
           $y_t\gets g_{\mathrm{DF}}(z_t^{\mathrm{DF}})$
    \STATE Set critic input:
           $x_t^V\gets x_t^{\mathrm{DF}}$

\ELSIF{$A=\textsc{Dual Memory}$}
    \STATE Update environmental memory
           $s_{o,t}^{\mathrm{DM}}$ using $e_{o,t}$
    \STATE Update interruption memory
           $s_{u,t}^{\mathrm{DM}}$ using $e_{u,t}$
    \STATE Concatenate hidden states:
           $\bar z_t^{\mathrm{DM}}\gets
           [z_{o,t}^{\mathrm{DM}};z_{u,t}^{\mathrm{DM}}]$
    \STATE Fuse memories:
           $y_t\gets g_{\mathrm{DM}}(\bar z_t^{\mathrm{DM}})$
    \STATE Set critic input:
           $x_t^V\gets[e_{o,t};e_{u,t}]$

\ELSIF{$A=\textsc{Gated}$}
    \STATE Compute scale and shift: $\alpha_t\gets f_\alpha(e_{u,t})$, $\delta_t\gets f_\delta(e_{u,t})$
    \STATE Modulate environmental features:
           $\widetilde e_{o,t}\gets
           \alpha_t\odot e_{o,t}+\delta_t$
    \STATE Update gated memory $s_t^{\mathrm{G}}$
           using $\widetilde e_{o,t}$
    \STATE Extract policy features:
           $y_t\gets g_{\mathrm{G}}(z_t^{\mathrm{G}})$
    \STATE Set critic input:
           $x_t^V\gets\widetilde e_{o,t}$
\ENDIF

\STATE Evaluate nominal policy:
       $\pi_{\phi,t}(\cdot)\gets\pi_\phi(\cdot\mid y_t)$
\STATE Select nominal proposal:
       $a_t\sim\pi_{\phi,t}(\cdot)$
\STATE Update critic memory $s_t^V$ using $x_t^V$
\STATE Estimate value:
       $V_{\psi,t}\gets V_\psi(g_V(z_t^V))$

\ENSURE $a_t$, $\pi_{\phi,t}(\cdot)$, $V_{\psi,t}$,
        and updated recurrent states
\end{algorithmic}
\end{algorithm}

\subsection{Interruption-Aware RL Training}

The proposed architectures must experience interruptions during training to learn how the interruption input changes at takeover and handback, and how recurrent memory should evolve between these events. Training therefore introduces randomised interruptions and implements a masked PPO actor update, as summarised in Algorithm~\ref{alg:2_interruption_aware_training}. For a fair comparison, the recurrent baseline is trained under the same conditions.

\begin{algorithm}[!t]
\caption{Interruption-aware recurrent PPO training}
\label{alg:2_interruption_aware_training}
\footnotesize
\begin{algorithmic}[1]
\REQUIRE Actor $\pi_\phi$, critic $V_\psi$, architecture $A$, interruption probability $p_{\mathrm{int}}(\kappa)$, override-sampling distributions, and PPO/GAE hyperparameters

\STATE Initialise network parameters and environment states

\FOR{each fixed-length rollout}
    \STATE Clear the rollout buffer

    \FOR{each environment step $t$}
        \IF{a new episode starts}
            \STATE Reset its recurrent states and action history
            \IF{selected with probability $p_{\mathrm{int}}(\kappa)$}
                \STATE Sample onset, duration,                       and supervisor to schedule override
            \ELSE
                \STATE Schedule nominal execution only
            \ENDIF
        \ENDIF

        \STATE Set previous-transition flag:
               $\lambda_t\gets\mathbb{I}(m_{t-1}>0)$
        \STATE Run Algorithm~\ref{alg:1_architecture_forward}
        \STATE Obtain nominal proposal $a_t$, value estimate $\widehat v_t$, and new states
        \STATE Determine the current action source $m_t$
               from the episode schedule

        \IF{$m_t=0$}
            \STATE Execute the nominal proposal:
                   $\bar a_t\gets a_t$
        \ELSE
            \STATE Obtain supervisor action $a_t^{(m_t)}$
            \STATE Execute the supervisor action:
                   $\bar a_t\gets a_t^{(m_t)}$
        \ENDIF

        \STATE Apply $\bar a_t$ and observe
               $r_t$, $o_{t+1}$, and episode termination
        \STATE Set actor mask:
               $w_t\gets\mathbb{I}(m_t=0)$
        \STATE Store the transition, nominal-action
               log-probability, $\widehat v_t$, and $w_t$
        \STATE Carry the updated recurrent states forward
    \ENDFOR

    \STATE Compute GAE advantages $\widehat A_t$
           and return targets $\widehat G_t$
    \STATE Mask actor advantages:
           $\widetilde A_t\gets w_t\widehat A_t$

    \FOR{each PPO epoch and recurrent minibatch}
        \STATE Reconstruct the ordered recurrent sequences
               using the stored episode-start flags
        \STATE Re-evaluate the actor and critic
               along each sequence
        \STATE Form PPO ratios using the stored
               nominal-action log-probabilities
        \STATE Compute the masked clipped actor objective
        \STATE Compute the unmasked value and entropy terms
        \STATE Update the network parameters using the PPO loss equation
    \ENDFOR
\ENDFOR

\ENSURE Trained actor and critic
\end{algorithmic}
\end{algorithm}

\subsubsection{Randomised interruption training}

The probability that an episode contains an interruption linearly increases throughout training, enabling the agent to first learn nominal deconfliction behaviour before progressively introducing interruptions. Each selected training episode contains a single bounded interruption whose onset, duration, and supervisor behaviour are sampled from predefined distributions. The specific training schedule and sampling distributions are specified in Section~\ref{section:4_evaluation}.

\subsubsection{Masked PPO actor updates}

During an interruption, the nominal actor continues to propose $a_t$ and update its recurrent state, but the supervisor-selected action $\bar a_t$ is executed. The resulting transition and reward cannot be attributed directly to $a_t$ and assigning its PPO advantage to that proposal would train the actor using an action it did not execute. The actor advantage is hence masked after GAE in \eqref{eq:27_masked_advantage}.
\begin{equation}
\widetilde A_t
=
\mathbb{I}(m_t=0)\widehat A_t.
\label{eq:27_masked_advantage}
\end{equation}

\noindent The masked advantage $\widetilde A_t$ is then used in the standard clipped PPO objective, so an interrupted step contributes no direct advantage-weighted policy gradient.

The critic and entropy terms remain unmasked. The critic is trained against returns from the executed-action trajectory, while the entropy term continues to regularise the policy at all visited histories. Actor and critic recurrent states are also unrolled through supervisor-executed steps. Masking therefore prevents direct credit assignment to unexecuted proposals without discarding the interrupted trajectories required to train recurrent memory and value estimation.

\section{Evaluation Methodology}
\label{section:4_evaluation}

\subsection{Simulation and Training Configurations}

Table~\ref{tab:3_simulation_training_configuration} reports the numerical settings used throughout training and experimentation, for the task described in Section~\ref{section:2_problem} and the RL architectures introduced in Section~\ref{section:3_proposal}.

\begin{table*}[!b]
\caption{Numerical Settings for Simulation and Training}
\label{tab:3_simulation_training_configuration}
\centering
\footnotesize
\setlength{\tabcolsep}{5pt}
\renewcommand{\arraystretch}{1.12}
\begin{tabularx}{\textwidth}{@{}
    >{\raggedright\arraybackslash}p{0.09\textwidth}
    >{\raggedright\arraybackslash}p{0.2\textwidth}
    >{\raggedright\arraybackslash}X@{}}
\hline
\textbf{Component} & \textbf{Parameter} & \textbf{Setting} \\
\hline

Environment
& Operating region
& $20\times20$\,m planar region; start and destination randomised by episode. \\

& Ego motion
& Constant ground speed of $1.5$\,m/s; normalised yaw-rate action $a_t\in[-1,1]$; maximum turn rate of $60^{\circ}$/s. \\

& Temporal resolution
& $\Delta t=0.1$\,s; maximum training-episode horizon of 500 steps. \\

& Goal and collision geometry
& Goal radius of $0.5$\,m; ego and intruder radii of $0.15$\,m. \\

\hline

Traffic
& Intruder count
& Training curriculum of $\{3,5,7,8,9,10\}$ intruders; ten intruders during evaluation. \\

& Intruder motion
& Constant ground speed of $0.5$\,m/s with reflection at the operating-region boundary. \\

& Heading perturbations
& Applied independently with probability $0.04$ per step; magnitude bounded by $0.7$\,rad. \\

\hline

Observation
& Traffic tracks
& $K=3$ nearest available tracks within an $8$\,m sensing range. \\

& Track refresh
& Common refresh period sampled uniformly from $\{2,\ldots,6\}$ timesteps; all tracks are refreshed together and held between updates with an increasing age indicator. \\

& Track noise and blanking
& Position-noise standard deviation of $0.05$\,m; traffic-track blanking probability of $0.15$ per step. \\

& Destination availability
& Destination-relative features are visible for five steps and blanked for the following 40 steps. \\
\hline

PPO
& Training budget
& $5\times10^{6}$ environment steps; learning rate of $3\times10^{-4}$. \\

& Rollout and optimisation
& Rollout length of 2048; minibatch size of 256; four optimisation epochs per rollout. \\

& Return estimation
& Discount factor of $\gamma=0.99$; GAE parameter $\lambda_{\mathrm{GAE}}=0.95$. \\

& PPO regularisation
& Clipping threshold of $0.2$; entropy coefficient of $0.01$; observation and reward normalisation enabled. \\
\hline

Networks
& Standard recurrent baseline
& Separate 64-unit actor and critic LSTMs; 64-unit post-LSTM layer. \\

& Proposed actors
& One 64-unit LSTM for dual-feature and gated actors; two 64-unit LSTMs for dual memory. \\

& Proposed critics
& One separate 64-unit critic LSTM for every architecture. \\
\hline

Reward
& Progress and completion
& $k_p=5$ and terminal goal bonus $B_g=50$. \\

& Separation and collision
& $k_c=2.5$, $d_{\mathrm{safe}}=1$\,m, and collision penalty $P_c=30$. \\

& Regularisation penalties
& Action smoothness $k_{\Delta a}=0.1$, time $c_{\mathrm{time}}=0.05$, and boundary contact $P_b=0.3$. \\
\hline

Curricula
& Traffic progression
& Advance after at least $75\%$ success over 50 episodes and $7.5\times10^{5}$ steps at the current traffic level. \\

& Interruption probability
& Linear increase from zero to $0.5$ over the first $2\times10^{6}$ training steps. \\

& Training interruption
& One segment per selected episode; onset is uniform over steps 20--200; duration is uniform over 5--20 steps; supervisor is sampled uniformly from the four-controller training mixture. \\

\hline
\end{tabularx}
\end{table*}

\subsection{Supervisor Behaviours}

Distinct supervisor behaviours are implemented to reflect safety-oriented, neutral, and adverse overrides. Only the first four behaviours are sampled uniformly during training, allowing for evaluation with an unknown supervisor.
\begin{itemize}
    \item \textbf{Handcrafted:} a heuristic supervisor that uses the same observation as the nominal policy. Intruder proximity is mapped to a set of turn commands that avoid collisions and track the observed goal bearing. This represents a simple on-board RTA system that uses the same observation as the nominal learning-based policy.

    \item \textbf{Simple oracular:} a privileged supervisor with access to the true ego and intruder kinematics. It samples a heading towards the goal and selects the closest yaw rate that remains outside a safety radius. This represents a ground-based RTA system with access to networked information.

    \item \textbf{Max-separation:} a safety-only oracular supervisor that turns away from the nearest intruder, with no goal term. This can represent a conservative remote pilot with access to additional networked information.

    \item \textbf{Antagonistic:} an inverted oracular planner that steers away from the goal and towards the nearest intruder, subject to a hard floor that avoids an immediate collision. It tests robustness to a strongly off-nominal supervisor and represents a malicious actor temporarily taking control.

    \item \textbf{Hold-heading:} a non-reactive supervisor that applies zero yaw rate and continues moving in straight line. This represents an interruption caused by temporary loss of the C2 link, during which the UAS continues moving in a straight line until the link is restored.
\end{itemize}

\subsection{Paired Experimental Design}

Matched simulation pairs are used to isolate interruptions from the effects of initial-condition and stochastic-traffic variability. For each environment seed, a \emph{reference arm} is executed entirely by the nominal policy, while an \emph{interrupted arm} uses the same initial conditions and environmental seed but includes an interruption, as illustrated in Fig.~\ref{fig:5_eval_arms}. Any differences can therefore be attributed to the interruption and its consequences.

\subsubsection{Scheduled interruptions}

The core study considers four supervisors with interruption onset $t^{\mathrm{on}}\in\{100,140\}$ and duration $d\in\{10,20\}$, each with 500 matched pairs.

\subsubsection{Hidden-state transplant}

\begin{figure}[!b]
\centering
\includegraphics[width=0.35\textwidth]{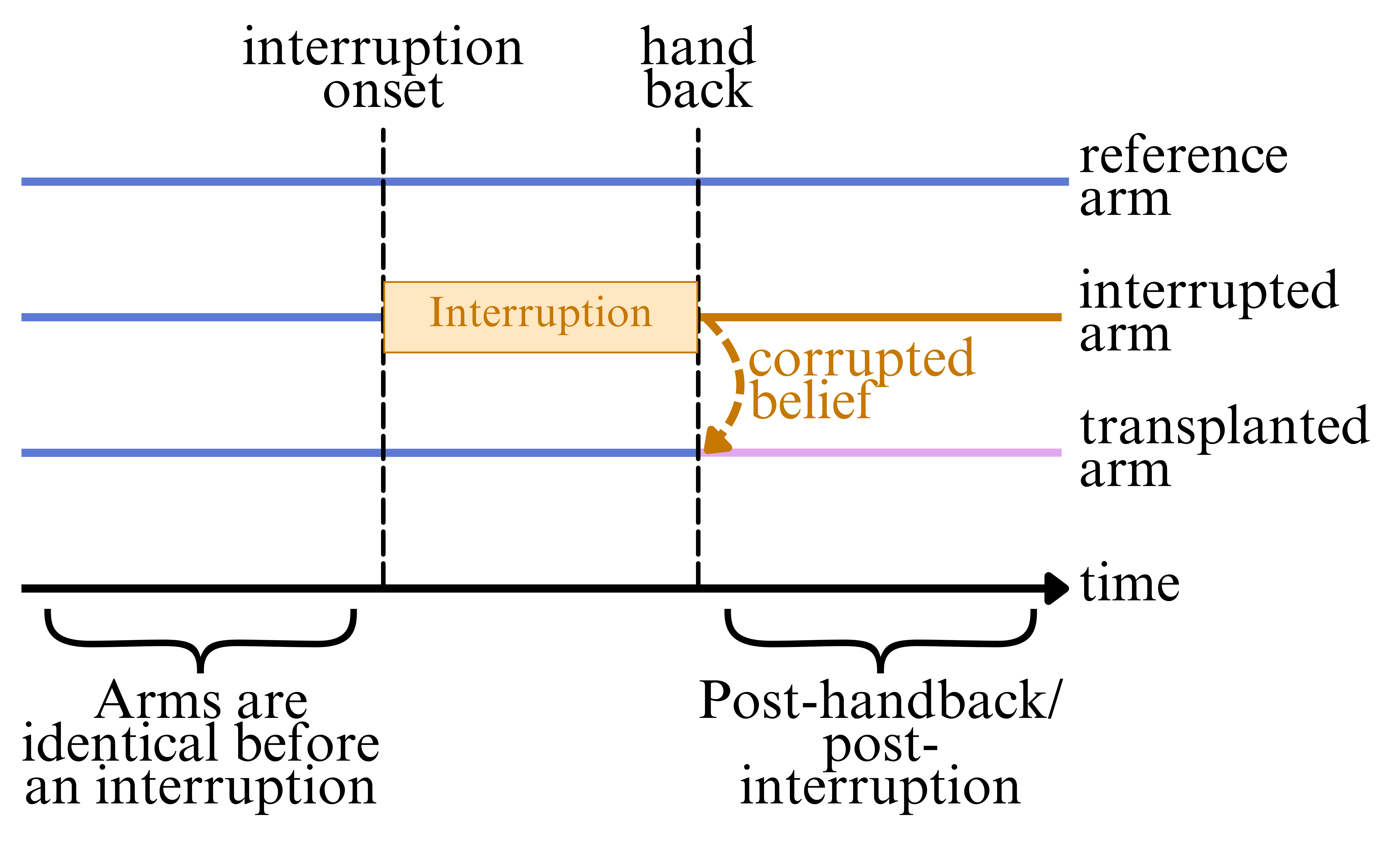}
\caption{Reference, interrupted and transplanted arms used to evaluate the impact of interruptions and interruption-aware recurrent architectures.}
\label{fig:5_eval_arms}
\end{figure}

A hidden-state transplant study is used to separate the effect of a degraded recurrent representation from the physical displacement caused by an interruption, considering interruptions with $t^{\mathrm{on}}\in\{60,100\}$ and $d=20$. A third arm is introduced that replays the reference arm's physical trajectory while injecting the hidden state of the interrupted arm at handback, as illustrated in Fig.~\ref{fig:5_eval_arms}.

\subsubsection{Multiple interruptions}

The effect of repeated interruptions is evaluated using the patterns $4\times5$, $2\times10$, $1\times20$, and $3\times10$, where the first value denotes the number of interruptions and the second their duration in steps.

\subsubsection{Naive interruption-handling}

Three naive solutions are applied by keeping the environmental observation untouched, freezing it at interruption onset, and masking it during the interruption. In each case, the effect of resetting the recurrent state after an interruption is also investigated.

\subsubsection{State-triggered interruptions}

State-triggered interruptions are evaluated using a max-separation supervisor that triggers when the nearest-intruder separation falls below $2$\,m and hands back after all intruders clear $3$\,m, subject to a maximum duration of 60 steps. A probabilistic trigger is first considered, whose per-step engagement probability increases as separation decreases, up to a maximum of $0.6$. A deterministic trigger is also implemented to consistently avoid collisions.

\subsubsection{Human-in-the-loop demonstration}

As a qualitative complement to the scripted supervisors, a human-in-the-loop demonstration is conducted through an interactive map-based interface based on the physics-informed PyBullet framework \cite{r48}. The operator observes the aircraft in real time, and may temporarily interrupt the agent to issue steering commands. This provides a complementary test with less structured supervisory behaviour.

\subsection{Hidden-State Probing}

To measure how much task-relevant information remains in recurrent memory, a linear ridge probe is fitted for each target variable using the actor's LSTM hidden state $z_t$, whereby successful decoding indicates that a variable is linearly recoverable from $z_t$ \cite{r12,r29}. Targets include ego kinematics, destination geometry, intruder ranges, the next nominal action, and past and future environmental states. Reference probes are trained on reference-arm states from 250 seeds and evaluated on 250 disjoint seeds, so scores reflect information that generalises across episodes. Since targets have different units and scales, each is standardised, and decodability is calculated separately per target using \eqref{eq:28_probe_decodability}.
\begin{equation}
\mathcal{D}=1-\mathrm{RMSE}^2,
\label{eq:28_probe_decodability}
\end{equation}

\noindent where $\mathcal{D}=1$ denotes perfect decoding. Scores are averaged across targets and evaluated over two windows. A sustained window covers steps 15-60 after handback and tests whether the degradation persists.

A probe may perform worse on interrupted-arm states because information is genuinely lost, corresponding to memory degradation. However, information may still be present but encoded differently after the interruption. To reduce sensitivity to such representational shifts, we leverage two additional metrics. The \emph{corrected gap} directly subtracts the shift from the reference probe, measured as the additional error incurred by applying the reference probe to interrupted-arm states. The \emph{within-arm gap} avoids the probe mismatch altogether by socring each arm using a probe fitted to that arm.

\subsection{Evaluation Metrics}

Experiments are evaluated using the following metrics, and uncertainty is reported as 95\% confidence intervals. In all cases, pairs in which an arm collides during the interruption or within 15 steps of handback are excluded to avoid reflecting collisions caused by poor supervisor performance, or a supervisor directly handing back authority before a collision.

\begin{itemize}

    \item \textbf{Post-handback collision rate:} the proportion of pairs in which the interrupted arm collides after handback, but the reference arm does not.

    \item \textbf{Lost-success rate:} the proportion of pairs in which the interrupted arm fails, but the reference arm succeeds.
    
    \item \textbf{Collision and success rates:} the unconditional proportion of episodes that collide or reach the destination.

    \item \textbf{Belief degradation:} the reduction in variable decodability between the reference and interrupted recurrent states, using the corrected or within-arm gap defined above.
\end{itemize}

\section{Results and Discussion}
\label{section:5_results}

\subsection{Recurrent Memory is Required for the Simulated Task}

\begin{figure}[!b]
\centering
\includegraphics[width=0.48\textwidth]{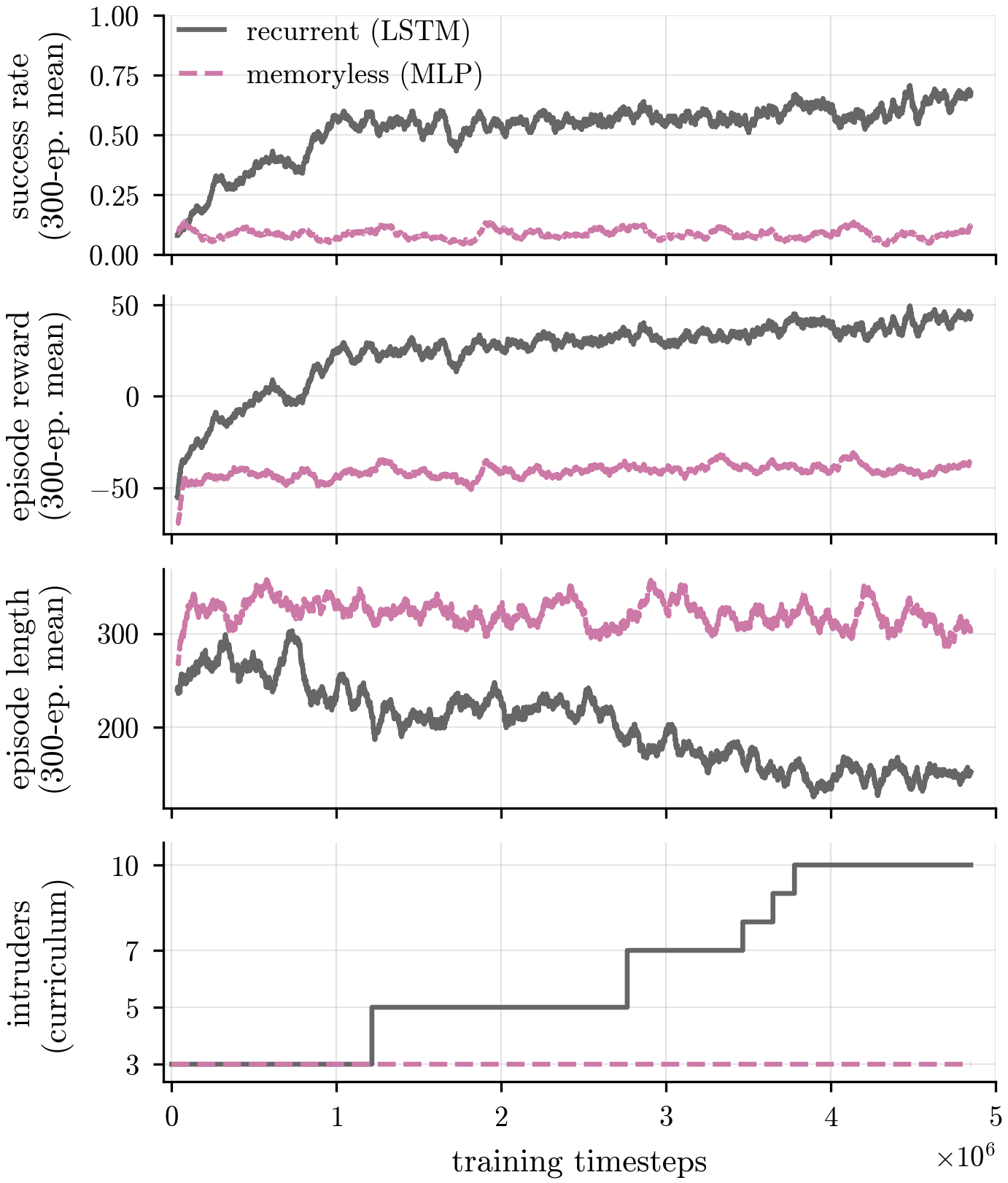}
\caption{Training curves of memoryless (pink) and recurrent (gray) PPO agents for the simulated UAS navigation task.}
\label{fig:6_need_for_recurrent_mem}
\end{figure}

Fig.~\ref{fig:6_need_for_recurrent_mem} compares the training curve of a memoryless multilayer perceptron (MLP) PPO agent \cite{r38} against the LSTM-based recurrent PPO baseline. Under the same training curriculum, the memoryless policy plateaus at a success rate of approximately 10\% with three intruders and never reaches the 75\% success threshold needed to progress the intruder curriculum. Conversely, the recurrent policy exceeds this threshold at every curriculum stage and progresses to the final ten-intruder setting. Its episode length also decreases as training progresses, consistent with a policy that generates increasingly efficient UAS trajectories. This confirms that recurrent memory is essential for effective navigation in the simulated task.

\subsection{Recurrent Memory Learns Task-Relevant Information}

\begin{figure}[!b]
\centering
\includegraphics[width=0.47\textwidth]{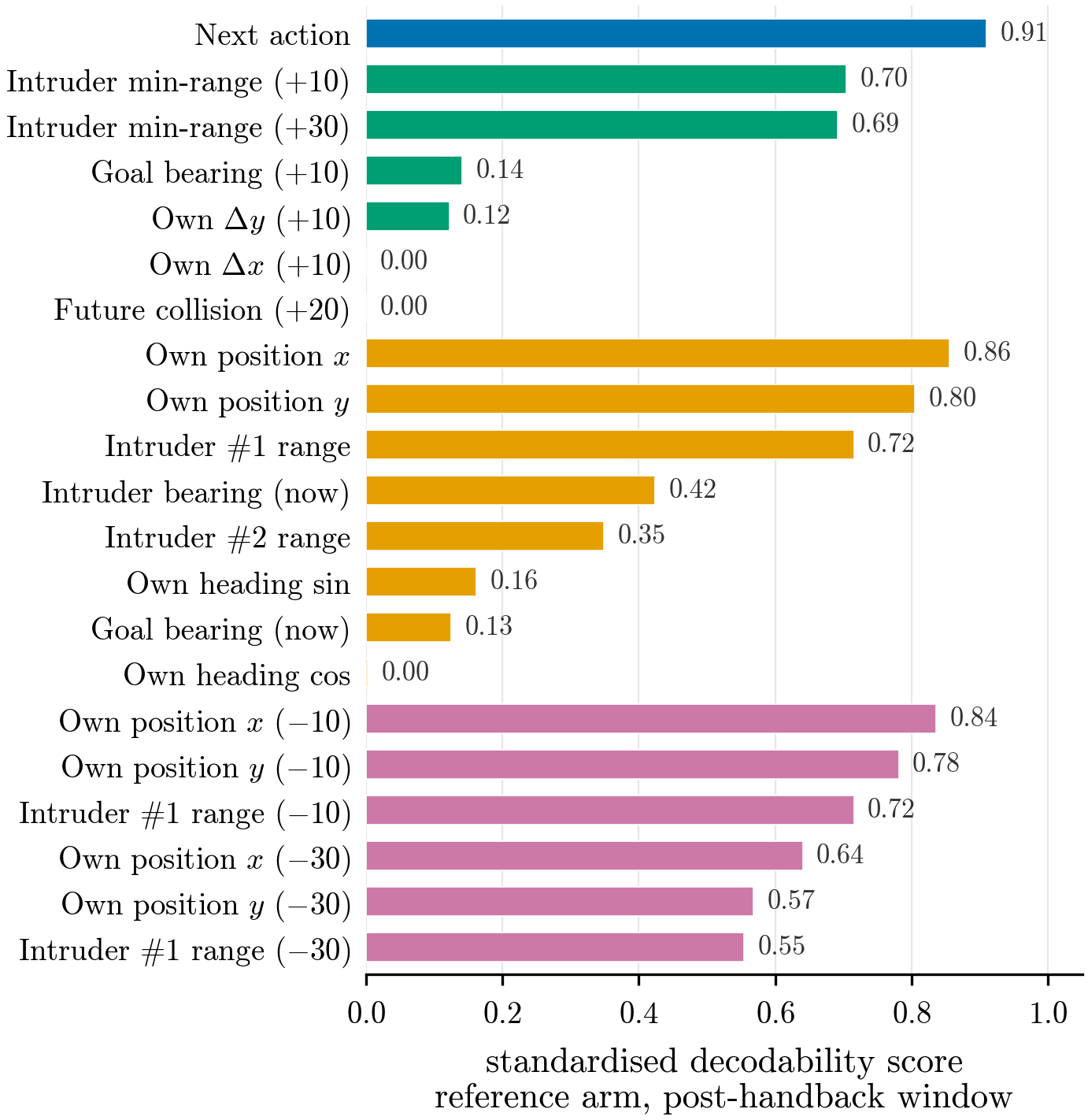}
\caption{Absolute decodability score of the reference arm hidden states for different task-relevant variables, averaged across a 45-step window.}
\label{fig:7_recurrent_decodability}
\end{figure}

Linear probes show that the actor's LSTM hidden state contains information relevant to the navigation task, as summarised in Fig.~\ref{fig:7_recurrent_decodability}. The current ego position ($\mathcal{D}=0.80$-$0.86$) and nearest-intruder range ($\mathcal{D}=0.72$) are strongly decodable from recurrent memory. The next action achieves the highest score ($\mathcal{D}=0.91$), since the policy head selects its action from the same hidden representation. The recurrent state also retains information that is no longer directly available in the current observation. Ego position and nearest-intruder range from 10 steps earlier remain strongly decodable ($\mathcal{D}=0.72$-$0.84$), with decodability decreasing to $0.55$-$0.64$ at a 30-step lag as expected for a gradually fading memory trace. Moreover, the future intruder range is decodable both 10 and 30 steps ahead ($\mathcal{D}=0.70$ and $0.69$). This suggests that the hidden state contains information about the evolving encounter geometry from which near-term separation can be predicted. These results therefore support interpreting the hidden state as a learned, task-dependent information state containing current, historical, and short-horizon predictive information. 

\begin{figure}[!b]
\centering
\includegraphics[width=0.47\textwidth]{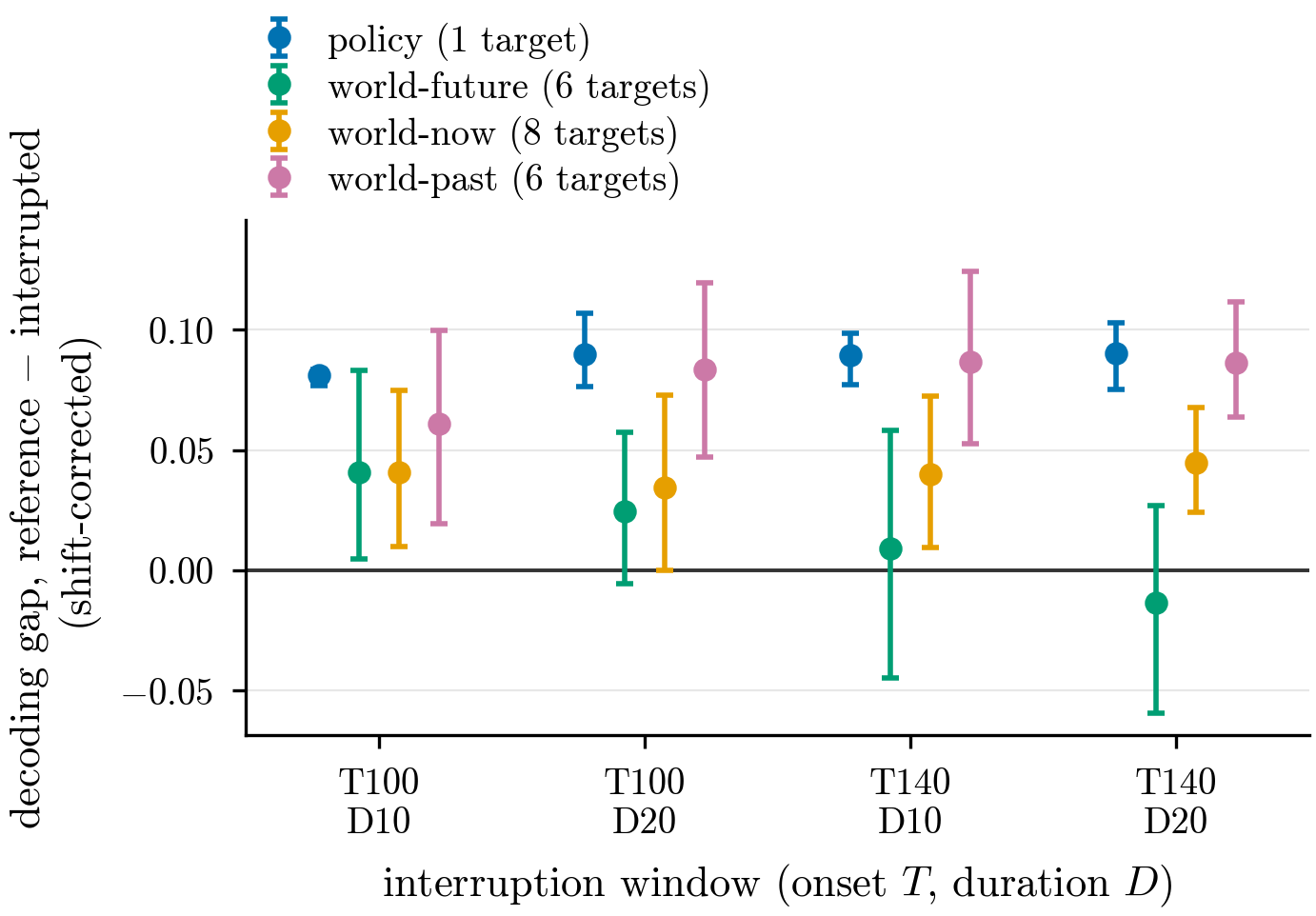}
\caption{Memory degradation caused by interruptions, measured as a reduction in decodability of task-relevant variables from the recurrent hidden state averaged across a 45-step post-handback window.}
\label{fig:8_recurrent_corruption}
\end{figure}

\subsection{Interruptions Reduce Task-Variable Decodability}

Fig.~\ref{fig:8_recurrent_corruption} compares the post-handback decodability gap between matched reference and interrupted arms for different task-relevant variables, and for different interruption onsets and durations. A positive value indicates that less information remains linearly recoverable after an interruption. 

The largest and most consistent corrected gaps occur for policy-readout and past-world variables, whereas current-world variables generally exhibit smaller gaps. This pattern is consistent with new observations helping to reconstruct current quantities, while historical quantities remain dependent on retained recurrent context. Future-world gaps are smaller and more variable, particularly for interruptions with later onsets. At a later handback, the remaining mission horizon is short and targets often collapse toward the present, reducing the impact of an interruption. Overall, the predominantly positive gaps confirm that interruptions reduce the task-relevant information recoverable from recurrent memory after handback.

\subsection{Interruptions Degrade Post-Handback Performance}

\begin{figure}[!b]
\centering
\includegraphics[width=0.47\textwidth]{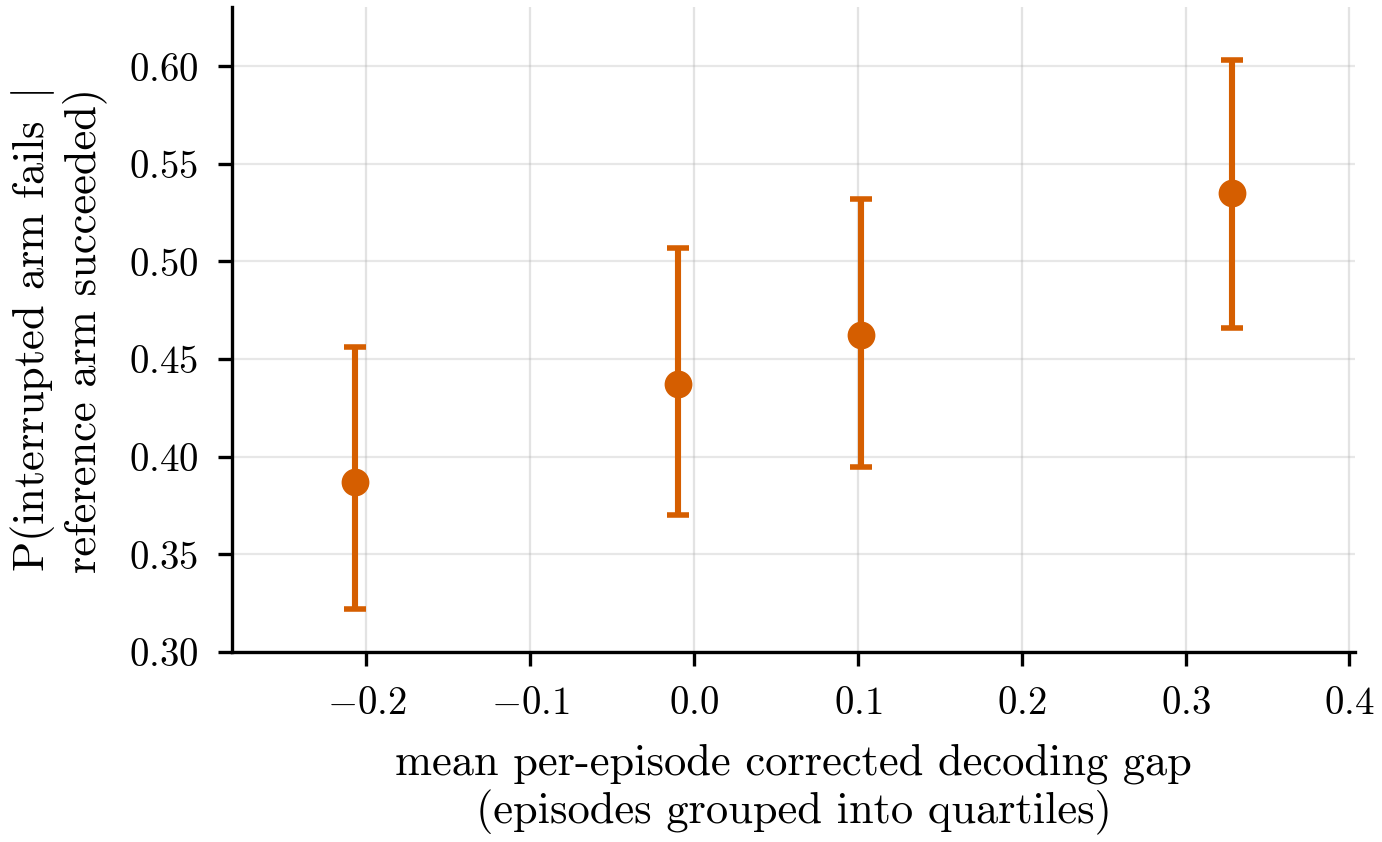}
\caption{Lost-success rate for episodes with different corrected decoding gaps, confirming that greater memory degradation results in poorer post-handback task performance.}
\label{fig:9_corrupted_mem_degrades_perf}
\end{figure}

At the episode level, larger decoding gaps are associated with poorer post-handback performance. Fig.~\ref{fig:9_corrupted_mem_degrades_perf} groups episodes into quartiles according to their corrected decoding gap and the lost-success rate increases monotonically from 39\% in the lowest quartile to 54\% in the highest. Notably, this association remains after controlling for displacement, heading error, and nearest-intruder distance at handback, confirming that reduced decodability is associated with interruption-induced mission failures beyond what can be explained by physical deviations at handback.

The transplant study further isolates the contribution of the recurrent state. At handback, the interrupted arm's actor recurrent state, comprising both hidden and cell states, is inserted into the matched reference arm while its physical state and current observation remain unchanged. This increases the post-handback collision rate by 4.6\% and suggests that the recurrent state produced during an interruption can independently increase post-handback collision risk even when the physical handback state is held constant. Nonetheless, recurrent memory is not the only reason why interruptions affect performance. In fact, both the modified aircraft trajectory and the altered recurrent state contribute to degraded post-handback performance.

\subsection{Action-Observation Feedback Sustains the Degradation}

\begin{figure}[!b]
\centering
\includegraphics[width=0.47\textwidth]{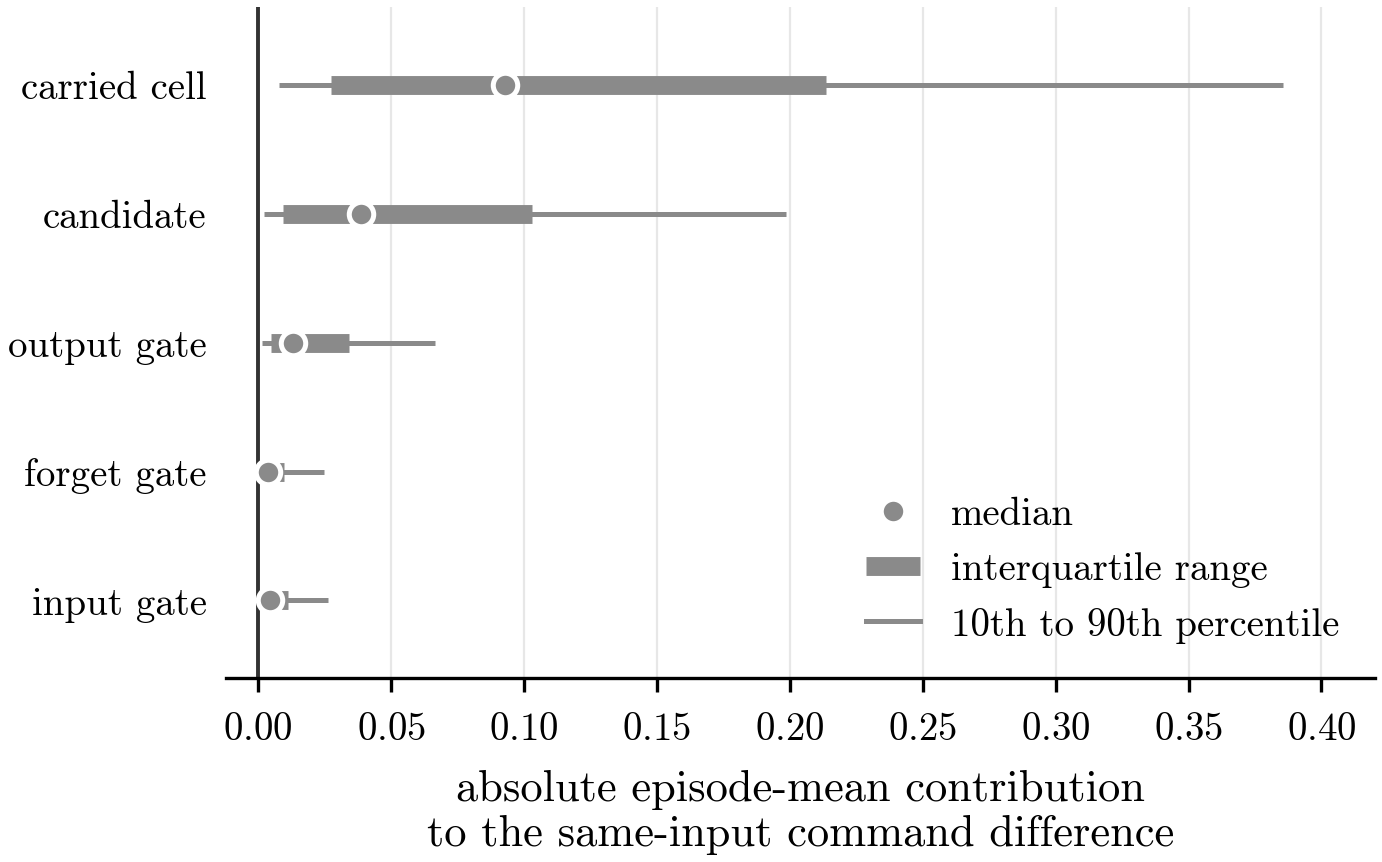}
\caption{Contribution of different signals to command differences, with carried cell content showing the largest contribution.}
\label{fig:10a_new_mech_carried_cell}
\end{figure}

\begin{figure}[!b]
\centering
\includegraphics[width=0.47\textwidth]{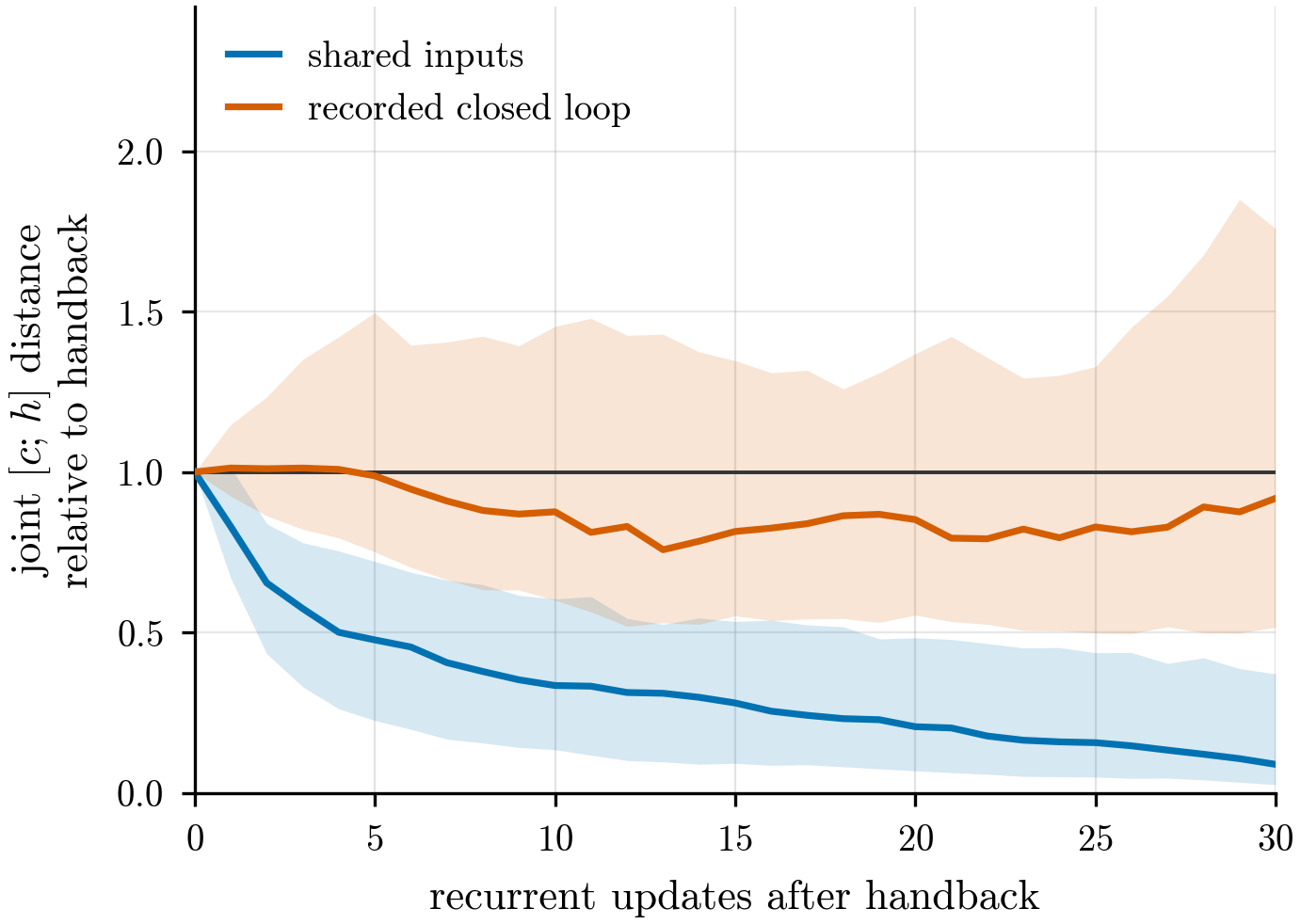}
\caption{Contraction of the reference and interrupted arm recurrent states after handback with shared inputs (blue) and closed-loop feedback (orange).}
\label{fig:10b_state_divergence}
\end{figure}

When supplying reference and interrupted arms identical observations after handback, the resulting command difference is illustrated in Fig.~\ref{fig:10a_new_mech_carried_cell}, decomposed into contributions from the LSTM carried cell, candidate update, and gate activation signals. Carried cell content shows the largest contribution, confirming that accumulated context influences post-handback commands. Moreover, under identical observations, recurrent-state differences normalised by their initial value at handback continue to decrease, as illustrated in Fig.~\ref{fig:10b_state_divergence}. The reduction is substantially weaker in actual closed-loop flights, where each arm follows its own trajectory and receives different observations. These results support a feedback hypothesis, whereby altered memory changes commands, while changed commands and interruption-induced displacement alter subsequent observations. Those observations in turn update recurrent memory, which shapes later commands in a closed feedback loop. This loop can continually generate new recurrent-state differences and prolong post-handback disruption, even as the original interruption-time differences diminish.

\subsection{Longer Contiguous Interruptions Are More Harmful}

\begin{figure}[!b]
\centering
\includegraphics[width=0.47\textwidth]{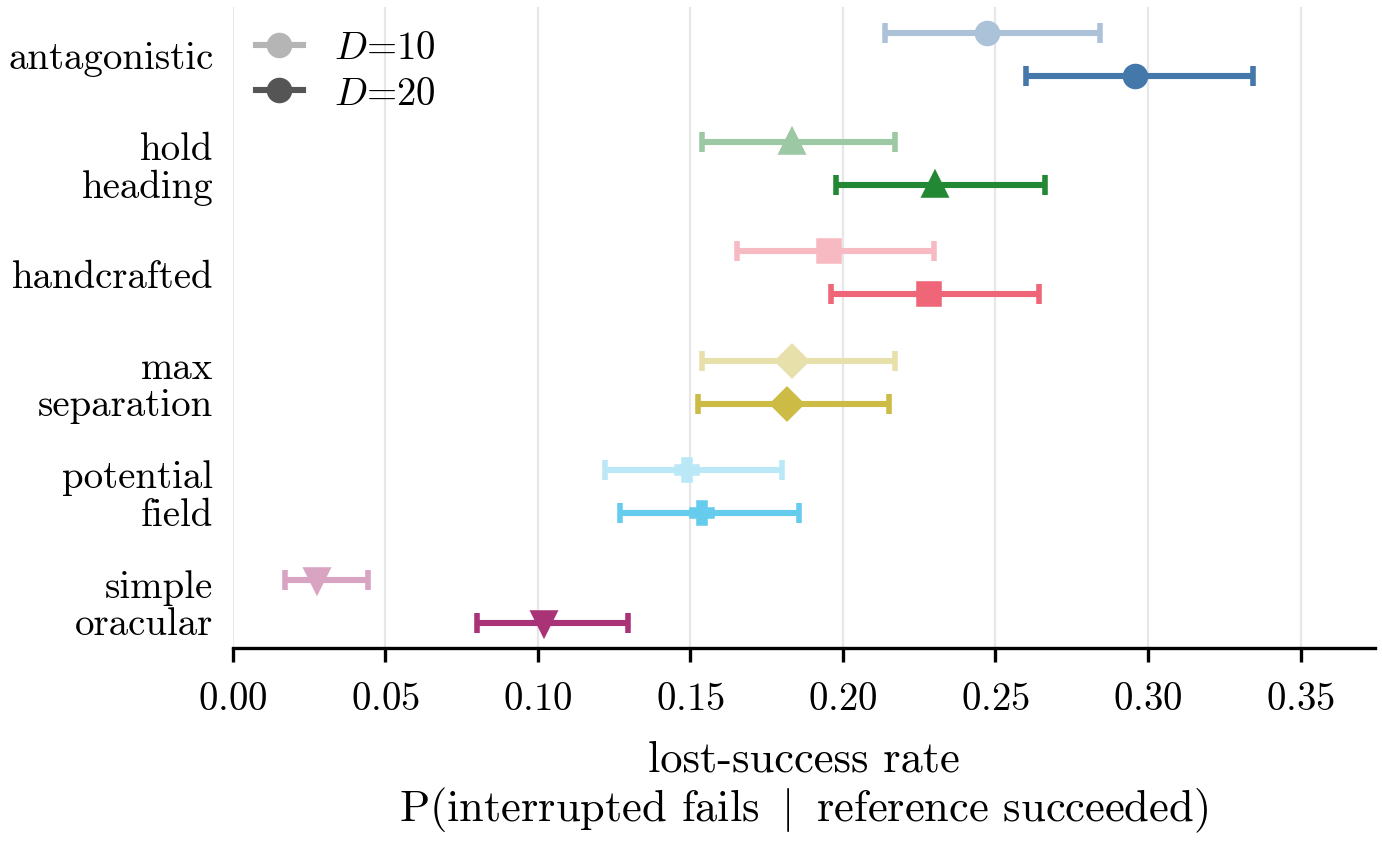}
\caption{Lost success rate for different supervisor behaviours and durations, showing that longer and less optimal supervisors are generally more harmful.}
\label{fig:11_supervisor_behaviours_and_durs}
\end{figure}

The effect of an interruption depends on supervisor behaviour and the duration for which it remains active, as shown in Fig.~\ref{fig:11_supervisor_behaviours_and_durs}. The antagonistic supervisor produces the highest lost-success rates, increasing from 25\% for a 10-step interruption to 30\% for a 20-step interruption. Hold-heading and handcrafted behaviours similarly increase from 18-20\% to 23\%. The potential-field and max-separation supervisors are less harmful, with rates of approximately 15\% and 18\%, respectively. The oracular planner produces the lowest lost-success rate at both durations. In general, longer and more off-nominal interruptions are more harmful.

\begin{figure}[!b]
\centering
\includegraphics[width=0.47\textwidth]{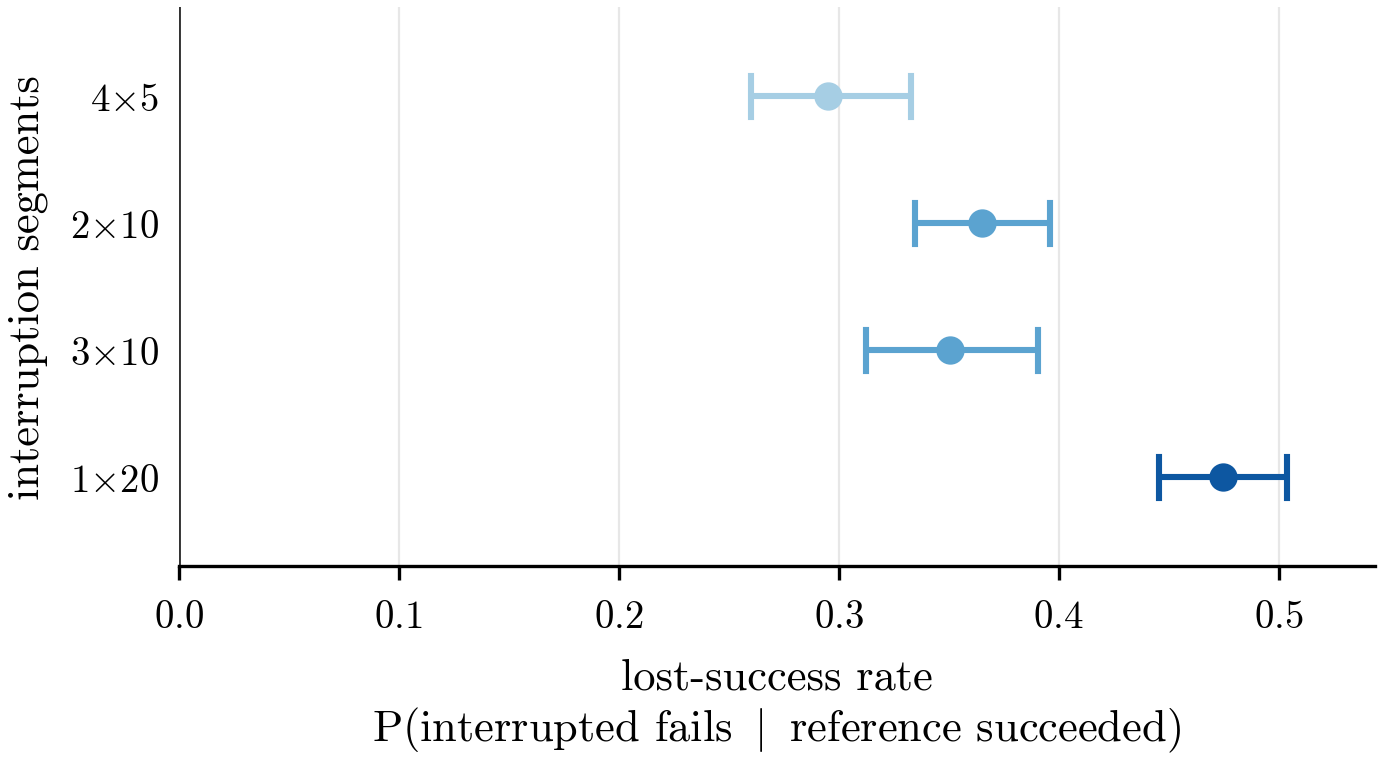}
\caption{Lost success rate for episodes with multiple interruption segments, demonstrating that longer contiguous interruptions are more harmful.}
\label{fig:12_multiple_interruptions}
\end{figure}

Fig.~\ref{fig:12_multiple_interruptions} shows that the lost-success rate increases from approximately 30\% for four five-step segments, to 37\% for two ten-step segments, and to 48\% for one continuous 20-step interruption. Conversely, three ten-step segments produce a lost-success rate of approximately 35\% despite increasing total interruption exposure to 30 steps. Mission harm is therefore more closely associated with the longest contiguous interruption than with the total interrupted time. This pattern is expected due to recurrent-state degradation accumulating during a sustained period of supervisor-generated transitions. Intermittent periods of nominal execution may allow the policy to incorporate new experience generated by the nominal policy and partially realign its recurrent state before the degradation becomes too severe.

\subsection{Naive Mitigations Are Insufficient }

\begin{figure}[!b]
\centering
\includegraphics[width=0.47\textwidth]{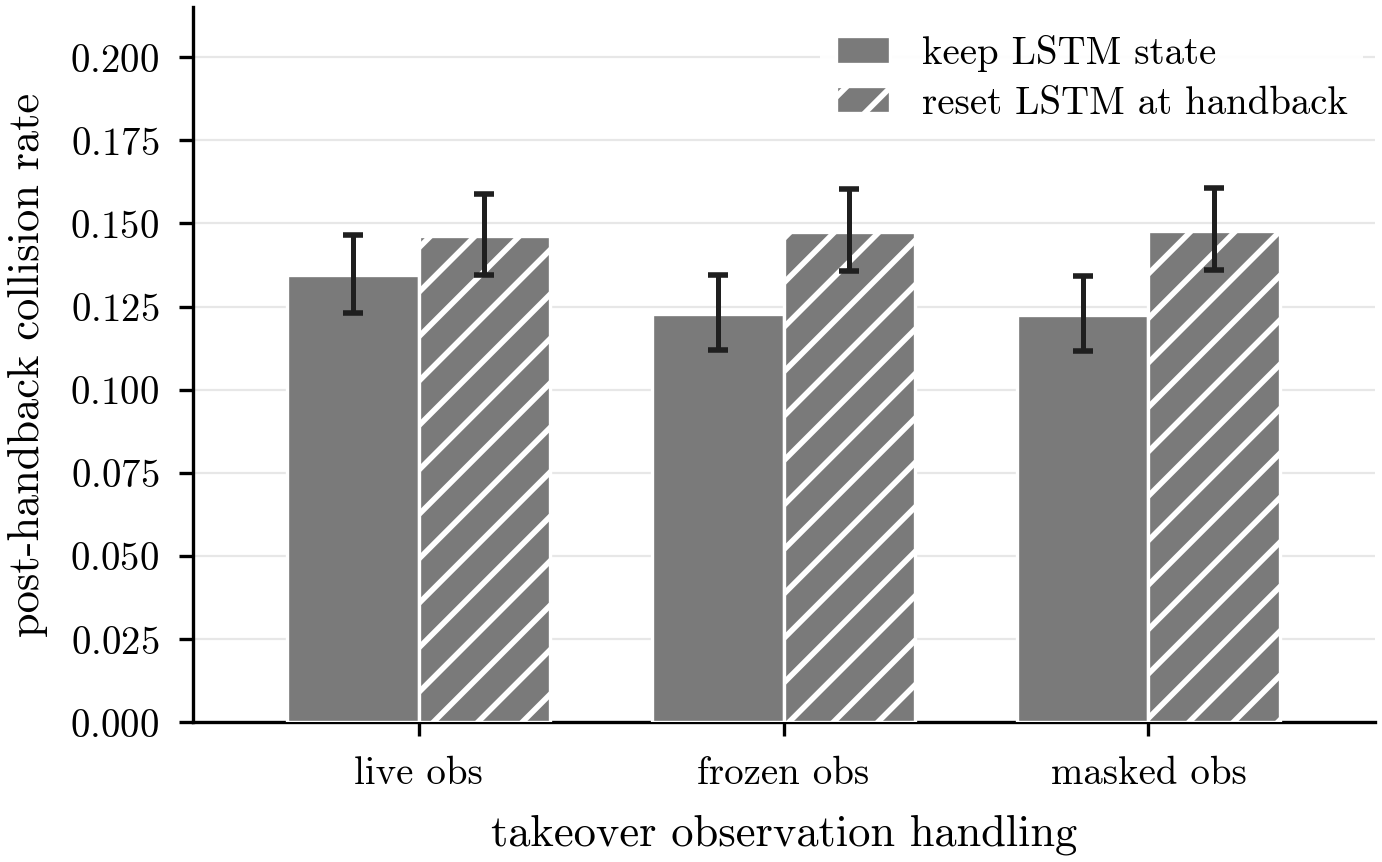}
\caption{Naive mitigations fail to improve post-handback performance.}
\label{fig:13_naive_fixes}
\end{figure}

Two simple mitigation strategies are investigated as naive solutions to external interruptions. Fig.~\ref{fig:13_naive_fixes} shows that resetting the recurrent state at handback worsens the post-handback collision rate since it discards useful memory along with interruption-associated context. Similarly, freezing or masking the observations during an interruption does not reliably improve post-handback performance.

\subsection{Interruption-Aware Architectures Preserve Task-Variable Decodability and Improve Post-Handback Performance}

\begin{figure}[!b]
\centering
\includegraphics[width=0.39\textwidth]{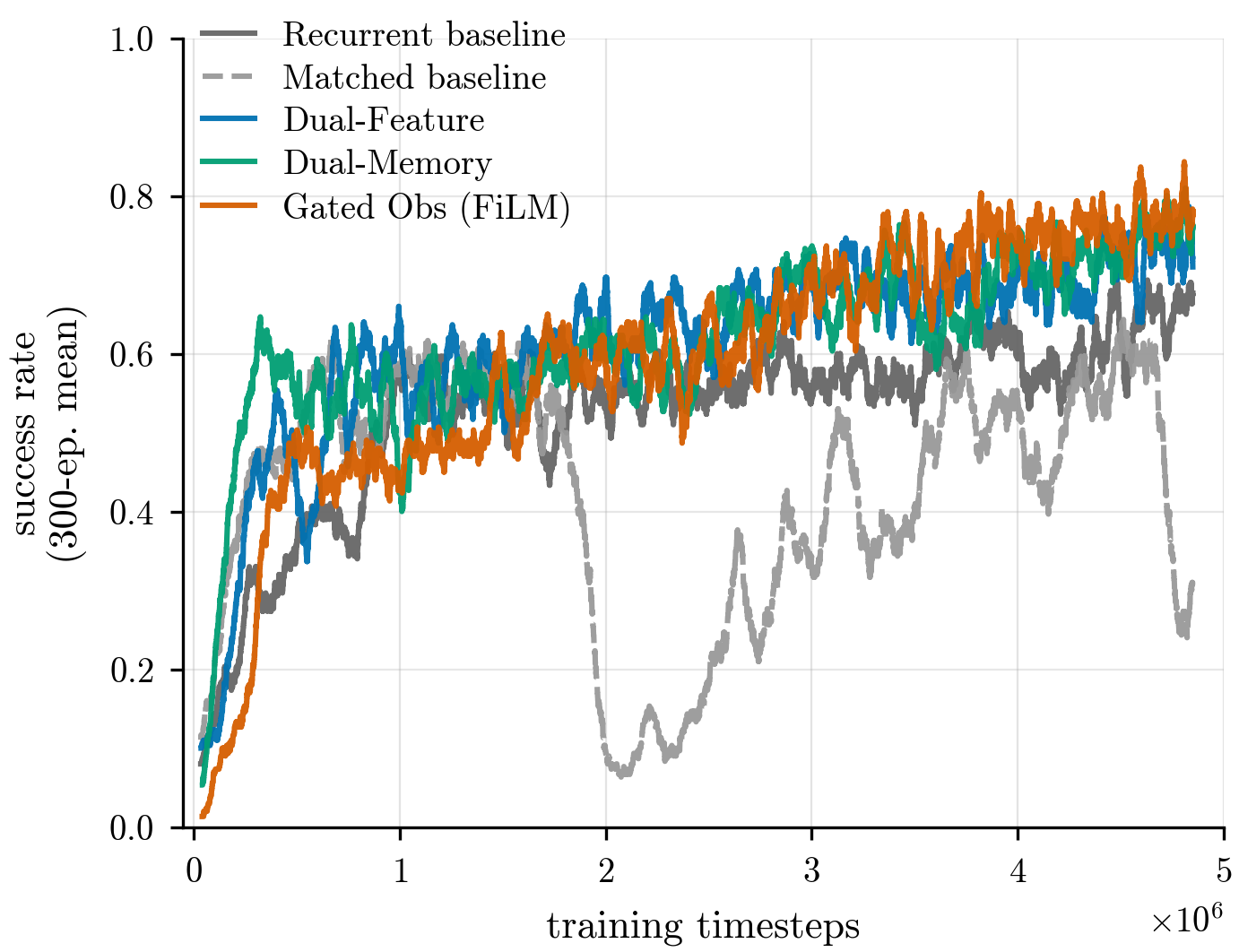}
\caption{Training curves showing convergence of interruption-aware and recurrent baseline policies.}
\label{fig:14_training_curves}
\end{figure}

Fig.~\ref{fig:14_training_curves} verifies that the proposed interruption-aware architectures learn the nominal navigation task without sacrificing interruption-free performance. All policies progress through the intruder curriculum, with the three interruption-aware architectures achieving comparable or higher success rates compared to the standard recurrent baseline. A capacity- and encoder-matched baseline is also included for reference.

\begin{figure}[!b]
\centering
\includegraphics[width=0.39\textwidth]{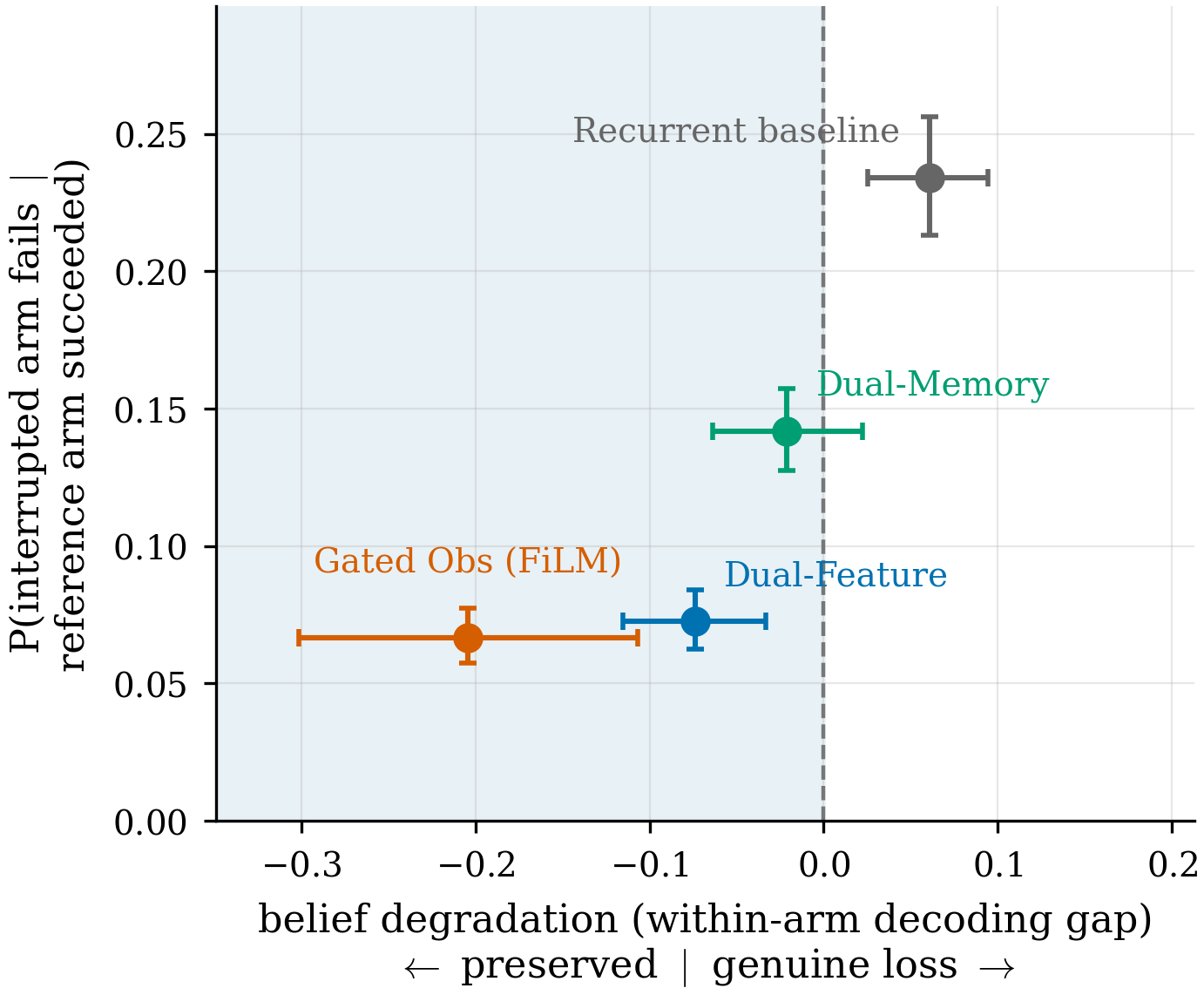}
\caption{Belief degradation against behavioural harm for each architecture, confirming that the interruption-aware architectures perform better than the recurrent baseline.}
\label{fig:15_arch_comparisons}
\end{figure}

\begin{figure}[!b]
\centering
\includegraphics[width=0.39\textwidth]{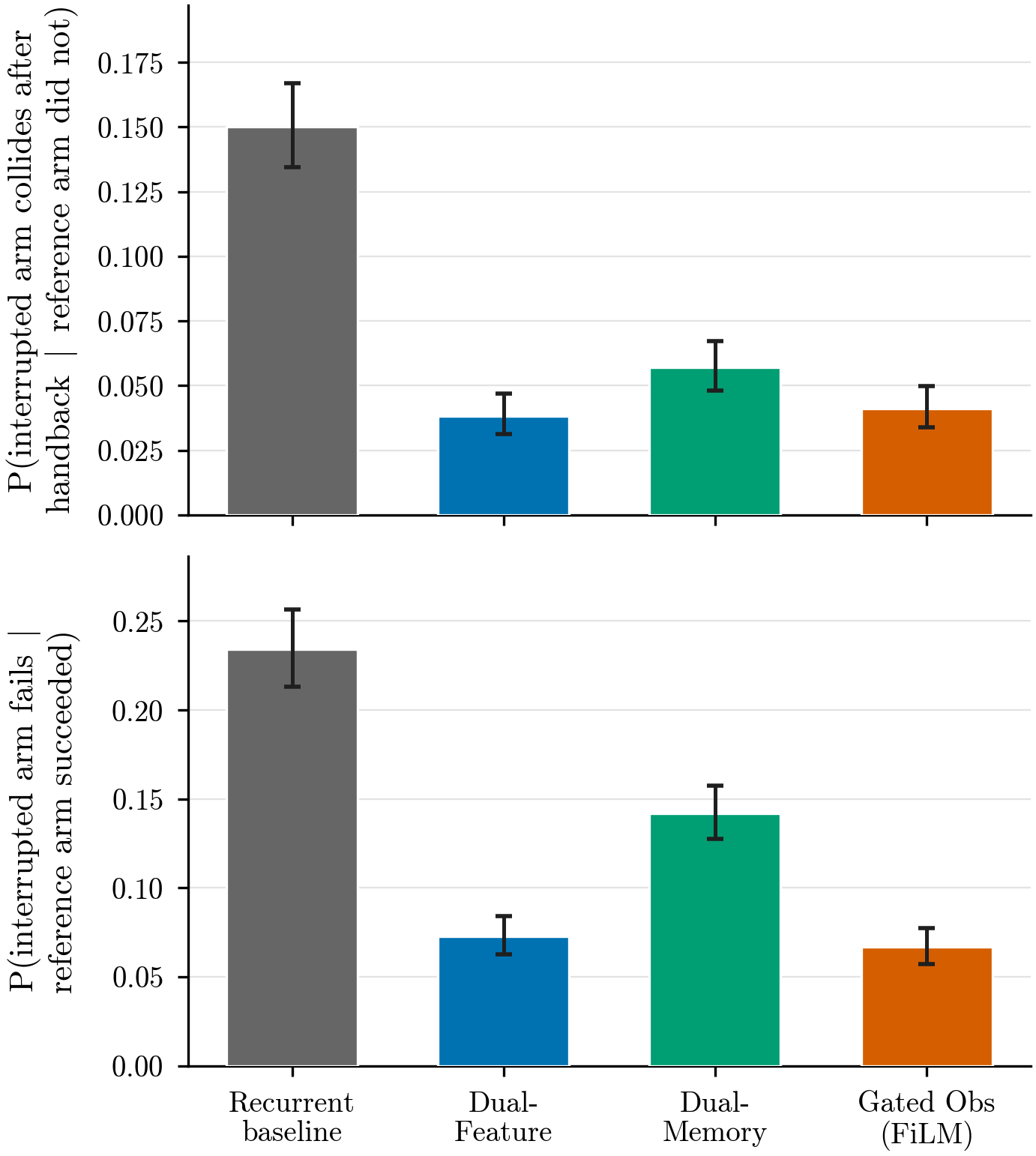}
\caption{Post-handback performance improvements with interruption-aware architectures.}
\label{fig:16_arch_results}
\end{figure}

Recurrent-memory degradation and interruption-induced mission harm for each architecture are summarised in Fig.~\ref{fig:15_arch_comparisons}. All three interruption-aware architectures mitigate the impact of interruptions, resulting in lower recurrent state degradation and improved post-handback performance. This is further confirmed in Fig.~\ref{fig:16_arch_results}, where the post-handback collision rate decreases from 0.15 for the recurrent baseline to 0.04-0.06 for the proposed architectures, and the lost-success rate decreases from 0.23 to 0.07-0.14. In the strongest case, these changes correspond to relative reductions of approximately 74\% in post-handback collisions and 69\% in interruption-induced lost successes. Moreover, memory-induced command differences were also observed to decay faster under the interruption-aware architectures. This shorter persistence limits the action–observation feedback that may otherwise prolong post-handback disruption. 

\begin{figure}[!t]
\centering
\includegraphics[width=0.47\textwidth]{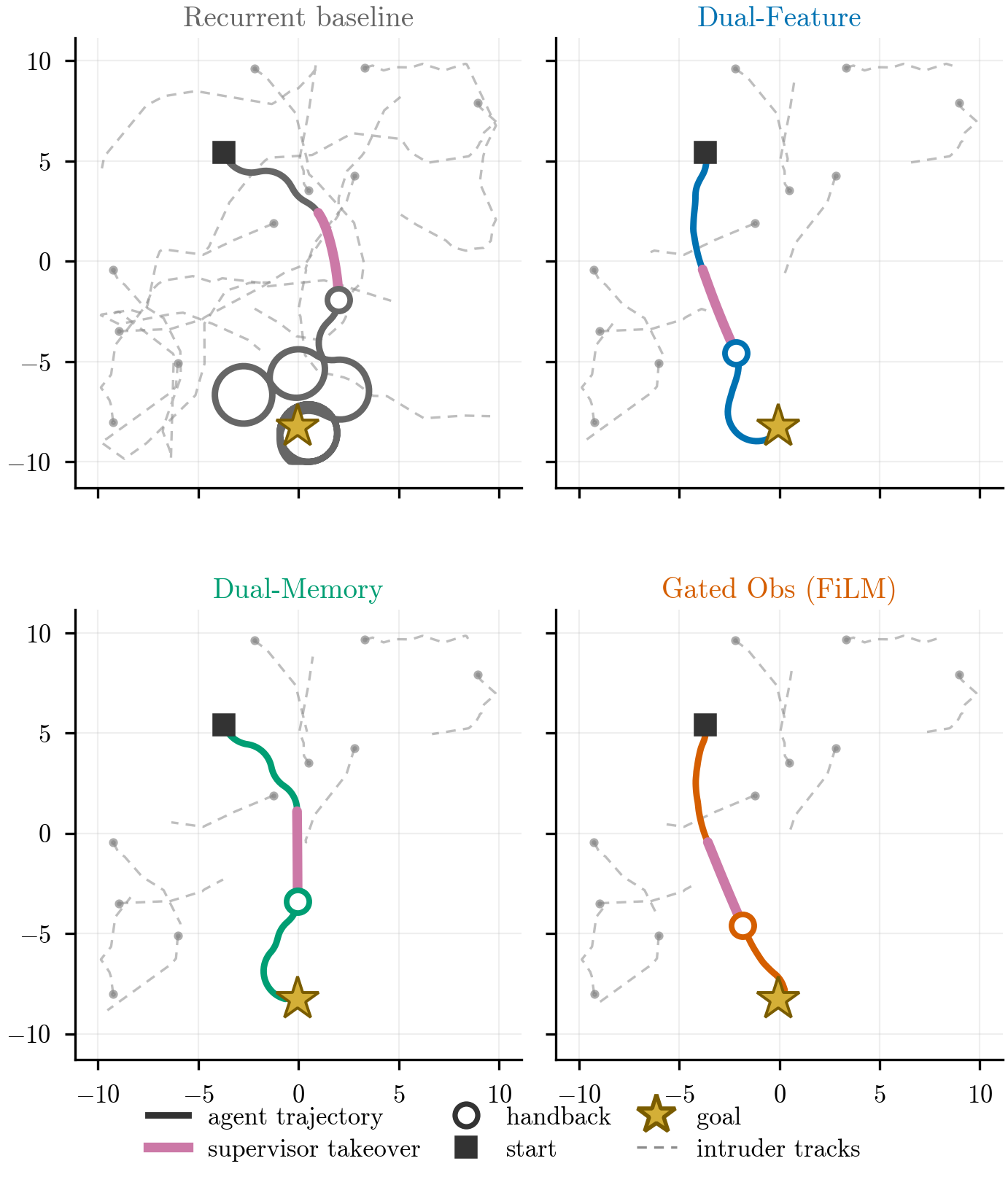}
\caption{Example trajectories generated by each architecture for the same environmental scenario.}
\label{fig:17_traj_example}
\end{figure}

Fig.~\ref{fig:17_traj_example} provides a qualitative example of the resulting behaviour. Under the same environment seed and scheduled interruption configuration, the recurrent baseline enters an unstable trajectory after handback and fails to reach the destination, whereas all three interruption-aware policies recover and complete the mission.

\subsection{Results Under State-Triggered and Human Interruptions}

\begin{figure}[!b]
\centering
\includegraphics[width=0.43\textwidth]{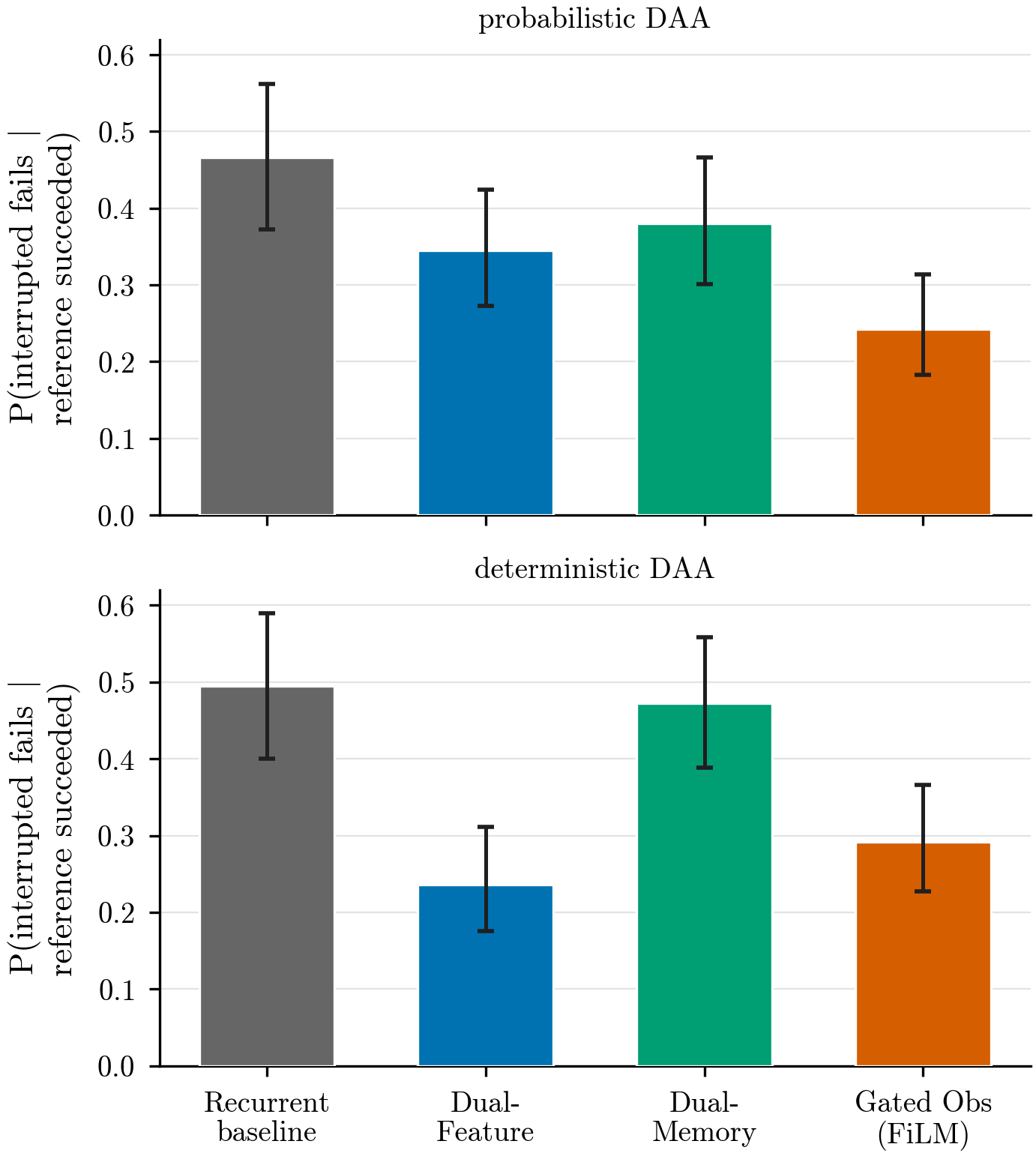}
\caption{Performance improvements of interruption-aware architectures with state-triggered interruptions.}
\label{fig:18_state_trig_interruptions}
\end{figure}

Operational safety interruptions normally respond to a developing traffic conflict. Fig.~\ref{fig:18_state_trig_interruptions} therefore evaluates the state-triggered supervisory scheme described in Section~\ref{section:4_evaluation}. Under probabilistic triggering, the lost-success rate is approximately 0.47 for the recurrent baseline, compared with 0.24-0.38 for the interruption-aware architectures. The gated architecture performs best in this regime. Under deterministic triggering, the dual feature achieves the lowest rate at approximately 0.24, followed by the gated architecture at 0.29. The dual memory architecture reaches approximately 0.47 and provides little improvement over the baseline. Consistent state-triggered interruptions therefore do not necessarily reduce failures. Although the supervisor is designed to address collision risk, the resulting manoeuvres and handback may still prevent an otherwise successful mission from reaching its destination.

\begin{figure}[!b]
\centering
\includegraphics[width=0.4\textwidth]{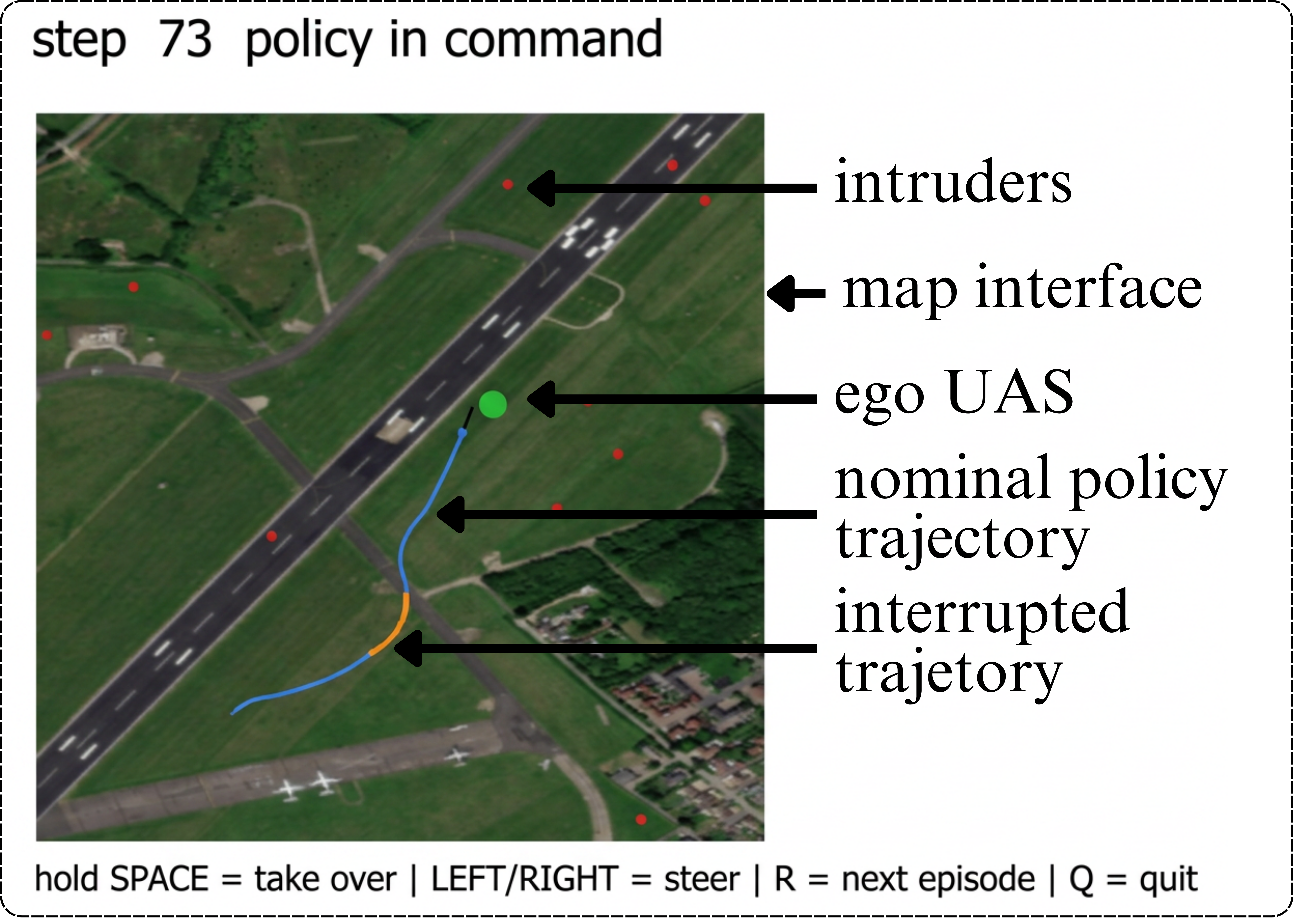}
\caption{Interface used for the human-in-the-loop demonstration, based on the physics-informed PyBullet simulation framework.}
\label{fig:19_human_interaction_gui}
\end{figure}

\begin{figure}[!b]
\centering
\includegraphics[width=0.43\textwidth]{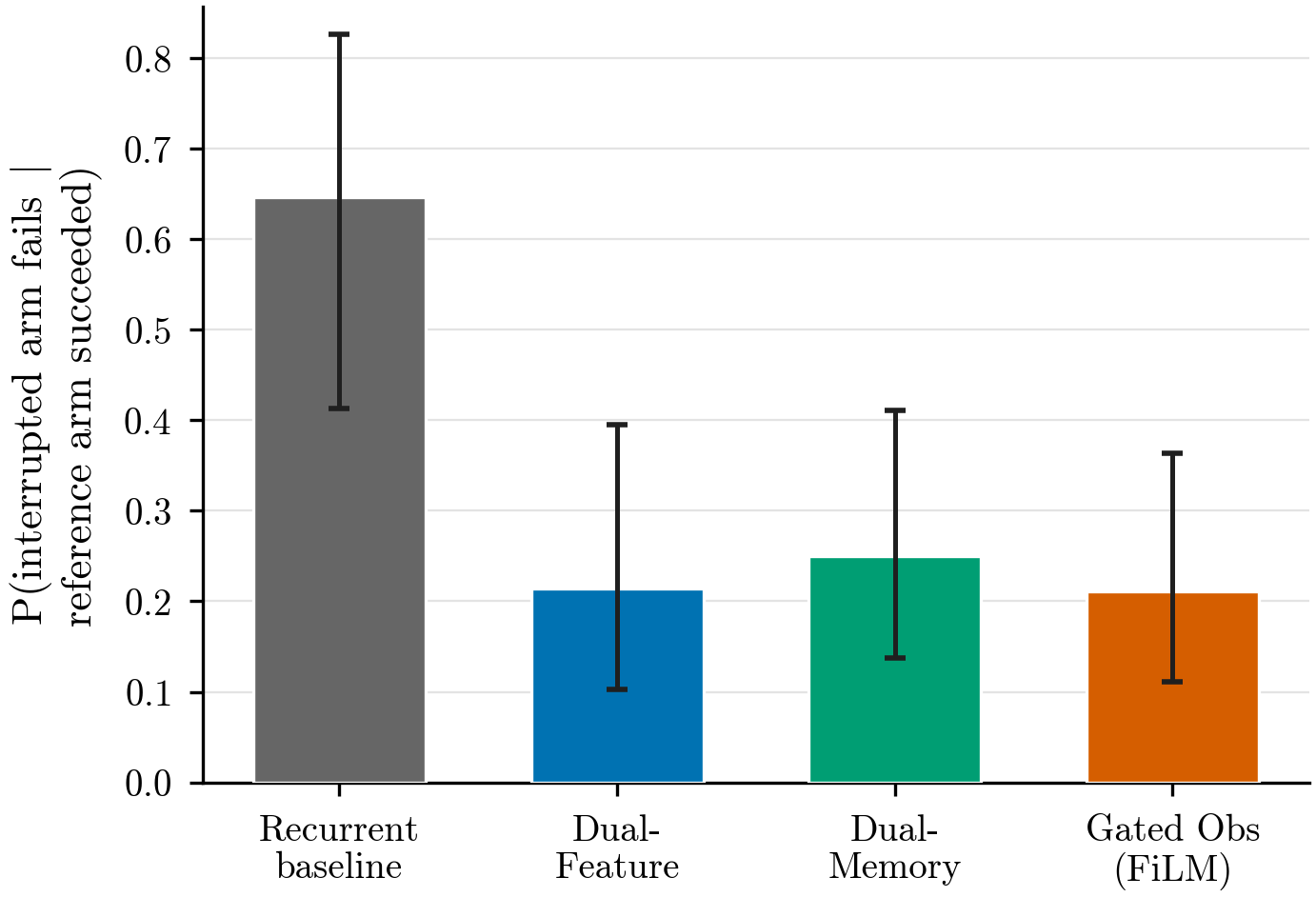}
\caption{Performance improvements of interruption-aware architectures with human operator-based interruptions.}
\label{fig:20_hitl_results}
\end{figure}

The human-in-the-loop experiment provides a complementary test using less structured supervisory behaviour. Through the interface shown in Fig.~\ref{fig:19_human_interaction_gui}, a human operator was instructed to intervene once per episode to avoid a collision or redirect the aircraft towards its destination. Forty human-interrupted episodes were collected for each architecture, with onset, duration, and steering commands selected by the operator. As shown in Fig.~\ref{fig:20_hitl_results}, the three interruption-aware architectures reduce the lost-success rate from approximately 0.65 under the recurrent baseline to 0.21-0.25. This demonstrates that the interruption-aware architectures also improve robustness to operator-generated manoeuvres. 

\subsection{Discussion and Operational Implications}

The results identify handback reliability as an additional safety requirement. Although the interruption may resolve an immediate hazard, it can degrade the policy's recurrent state and reduce task-variable decodability, especially for historical variables that cannot be reconstructed from new observations. This can increase collision risk and degrade post-handback performance. Notably, this vulnerability extends beyond the handback instant. Carried cell content continues to influence commands, while changed commands and supervisor-induced displacement alter subsequent observations. Those observations update the memory used for later decisions, sustaining a post-handback disruption feedback loop. Recovery therefore depends on how recurrent context is used and updated when the agent regains authority.

The risk of an interruption varies with supervisor behaviour and interruption duration. Safety- and goal-directed supervisors generally cause fewer induced failures, but even helpful interruptions can cause reference successes to fail. Longer contiguous overrides are particularly disruptive as they allow recurrent state degradation to accumulate. Resetting memory and masking or freezing observations during an interruption provide no consistent improvement. Conversely, all three proposed interruption-aware architectures preserve more task-relevant information and improve post-handback performance, including under unseen supervisor behaviours. Explicit override context enables the proposed architectures to recalibrate how carried recurrent memory influences commands after handback. This shorter persistence can limit the action-observation feedback loop that otherwise prolongs post-handback disruption. Among the proposed designs, dual feature and gated architectures achieve the best performance, while the dual memory architecture performs less strongly but still improves substantially over the baseline. Dual memory's weaker results may reflect its integration of interruption context at action selection, whereas the dual feature and gated architectures allow that context to directly shape how environmental memory is formed and updated.

\subsection{Limitations}

The principal experiments use simplified two-dimensional dynamics, sensing, traffic and supervisor behaviours. Moreover, state-triggered evaluations and the single-operator human study only provide preliminary evidence of broader applicability. Higher-fidelity evaluation should incorporate three-dimensional dynamics, realistic surveillance and communication delays, standards-aligned safety functions, and more extensive human-in-the-loop trials. Additionally, the probes used in this study measure how readily task variables can be recovered from the recurrent hidden state. Lower decodability, however, does not necessarily imply that information has been irreversibly lost. Future work should combine non-linear probes with carefully controlled interruptions to clarify what information remains available and better study how it influences post-handback decisions. Further training seeds can also be leveraged for more robust results. 

\section{Conclusion}
\label{section:6_conclusion}

This paper characterises temporary overrides as a handback reliability problem for recurrent UAS navigation under partial observability. The problem is formulated as a PO-MAMDP that distinguishes nominally proposed actions from externally modified actions. Controlled experiments show that interruptions can reduce task-variable decodability, cause otherwise successful missions to fail, and increase collision risk through changes in the recurrent state. Notably, harm depends on supervisor behaviour and the longest contiguous interruption. Moreover, degradation is likely sustained through a post-handback feedback loop in which carried recurrent context influences commands and subsequent experience renews recurrent-state differences after handback.

Three interruption-aware architectures are proposed to explicitly represent interruption context through separate feature streams, dedicated interruption memory, or gated environmental features. These reduce the persistence of interruption-associated memory effects on commands, limiting action-observation feedback that could prolong post-handback disruption. Under scheduled interruptions, the strongest evaluated design reduces post-handback collision harm by 74\% and interruption-induced lost successes by 69\% relative to the recurrent baseline, without degrading interruption-free performance. State-triggered and pilot-generated tests provide preliminary evidence that these benefits extend beyond scheduled supervisor behaviour.

These findings establish handback reliability as a distinct reliability concern for recurrent autonomy. Future work should assess robustness across training seeds and higher-fidelity environments, explore solutions that recover incorrect carried recurrent context, and explore the impact of interruptions on different learning algorithms and recurrent architectures. Moreover, future research should investigate the impact of interruptions in hierarchical and multi-agent navigation settings.

\section*{Acknowledgments}
The authors would like to thank the European Commission and the SESAR Joint Undertaking for the initiation and funding of the SAFIR-Ready project. The SAFIR-Ready project has received funding from the European Union’s Horizon Europe research and innovation programme under grant agreement No. 101114855.

\bibliographystyle{IEEEtran}
\bibliography{Bibliography}

\end{document}